\documentclass[letterpaper]{article} 
\usepackage{aaai2026}  
\usepackage{times}  
\usepackage{helvet}  
\usepackage{courier}  
\usepackage[hyphens]{url}  
\usepackage{graphicx} 
\usepackage{natbib}  
\usepackage{caption} 
\usepackage{algorithm}
\usepackage{algorithmic}

\newcommand{\ysq}{\textcolor[rgb]{0,0,0}}
\newcommand{\jy}{\textcolor[rgb]{0,0,0}}

\newcommand{\ryn}{\textcolor[rgb]{0,0,0}}

\newcommand{\rl}{\textcolor[rgb]{0,0,0}}

\newcommand{\zyj}{\textcolor[rgb]{0,0,0}}
\newcommand{\zyjn}{\textcolor[rgb]{0,0,0}}

\usepackage{pifont}

\newcommand{\ysqq}{\textcolor[rgb]{0,0,0}}
\newcommand{\zyja}{\textcolor[rgb]{0,0,0}}

\usepackage{amsmath}
\usepackage{amssymb}

\usepackage{algorithm}
\usepackage{algorithmic}
\usepackage[svgnames]{xcolor}
\usepackage{multirow}
\usepackage{makecell}

\usepackage{colortbl}

\usepackage{graphicx}
\usepackage{float}

\newcommand{\eg}{\textit{e.g.}}  
\newcommand{\ie}{\textit{i.e.}}

\usepackage{enumitem}

\PassOptionsToPackage{numbers, compress}{natbib}

\usepackage{amsmath}
\usepackage{caption}

\usepackage[utf8]{inputenc} 
\usepackage[T1]{fontenc}    
\usepackage{url}            
\usepackage{booktabs}       
\usepackage{amsfonts}       
\usepackage{nicefrac}       
\usepackage{microtype}      
\usepackage{xcolor}         

\usepackage{newfloat}
\usepackage{listings}
\DeclareCaptionStyle{ruled}{labelfont=normalfont,labelsep=colon,strut=off} 
\floatstyle{ruled}
\newfloat{listing}{tb}{lst}{}
\floatname{listing}{Listing}
\title{OpenScan: A Benchmark for Generalized Open-Vocabulary 3D Scene Understanding}
\author {
    Youjun Zhao\textsuperscript{\rm 1}, \ \ 
    Jiaying Lin\textsuperscript{\rm 1,2,}\thanks{\ Jiaying Lin and Rynson Lau are the corresponding authors.}, \ \ 
    Shuquan Ye\textsuperscript{\rm 1,3}, \ \ 
    Qianshi Pang\textsuperscript{\rm 4}, \ \ 
    Rynson W.H. Lau\textsuperscript{\rm 1,*}
}
\affiliations {
    \textsuperscript{\rm 1} City University of Hong Kong\\ 
    \textsuperscript{\rm 2} The Hong Kong University of Science and Technology\\ 
    \textsuperscript{\rm 3} The Chinese University of Hong Kong 
    \textsuperscript{\rm 4} South China University of Technology\\
}

\begin{document}

\maketitle

\begin{abstract}
Open-vocabulary 3D scene understanding (OV-3D) aims to localize and classify novel objects beyond the closed \zyjn{set of} object classes. However, existing approaches and benchmarks primarily focus on \ryn{the \ysq{open vocabulary} problem} within the context of object classes, which is insufficient \rl{in providing} a holistic evaluation to what extent a model understands the 3D scene. In this paper, we introduce a more challenging task called Generalized Open-Vocabulary 3D Scene Understanding (GOV-3D) to explore \ryn{the \ysq{open vocabulary} problem} beyond object classes. \ryn{It encompasses an open and diverse set of} \ysq{generalized knowledge, expressed as linguistic queries of fine-grained and object-specific attributes.} \ryn{To this end, we contribute} a new benchmark named \textit{OpenScan}, which consists of \ysq{3D object attributes across eight representative linguistic aspects}, \ryn{including \zyj{affordance}, \ryn{\zyj{property}, \ryn{and material}.}}
We further evaluate state-of-the-art OV-3D methods on \ryn{our} OpenScan benchmark and discover that these methods struggle to comprehend the abstract vocabularies \ysq{of the GOV-3D task, a challenge that cannot \ryn{be} addressed simply by scaling up object classes during training.} We highlight the limitations of existing methodologies and explore promising directions to overcome the identified shortcomings.
\end{abstract}

\vspace{-3.8mm}
\begin{links}
    \link{Code, Datasets, and Extended version}{https://youjunzhao.github.io/OpenScan/}
\end{links}
\vspace{-3.8mm}

\section{Introduction}
Open-vocabulary 3D scene understanding (OV-3D) involves recognizing \zyjn{object classes} \zyjn{that} are not \zyjn{included} in the training set.
It \ryn{is important to applications} like \ryn{autonomous driving~\cite{autodrive} and robotics~\cite{robotic}}.
Recently, vision-language models (VLMs), \textit{e.g.}, CLIP~\cite{clip}, have achieved significant progress by leveraging large-scale image-text datasets with semantically rich captions. The impressive capability of VLMs in capturing rich contexts between images and texts has inspired further exploration in \ryn{open-vocabulary tasks in both 2D\zyj{~\cite{gu2021open,regionclip}} and 3D\zyj{~\cite{takmaz2023openmask3d,peng2023openscene}} domains}.

\ysq{For AI systems, the capability to comprehend diverse linguistic aspects of object-related attributes and their association with corresponding \rl{objects is as important} as the identification of the objects themselves. Consequently, the field of open-vocabulary 3D scene understanding should ideally extend beyond specific object classes to encompass complex object-related attributes, \rl{such as affordance and material, articulated through natural languages}.}
\ysq{However, the generalization ability of existing OV-3D methods~\cite{peng2023openscene, takmaz2023openmask3d, yan2024maskclustering, sai3d,nguyen2023open3dis} \rl{to object-related} attributes has not been thoroughly and systematically explored.}
\zyj{Besides, evaluating the ability of an OV-3D model to recognize specific object attributes is difficult due to the shortage of large-scale OV-3D attribute benchmarks. Existing OV-3D benchmarks, such as ScanNet~\cite{scannet} and ScanNet200~\cite{scannet200}, \ryn{primarily focus} on object classes, and do not \rl{include annotations of object attributes for evaluating} the generalized ability of OV-3D methods.}

\begin{figure}[tb] \centering
    \includegraphics[width=0.46\textwidth]{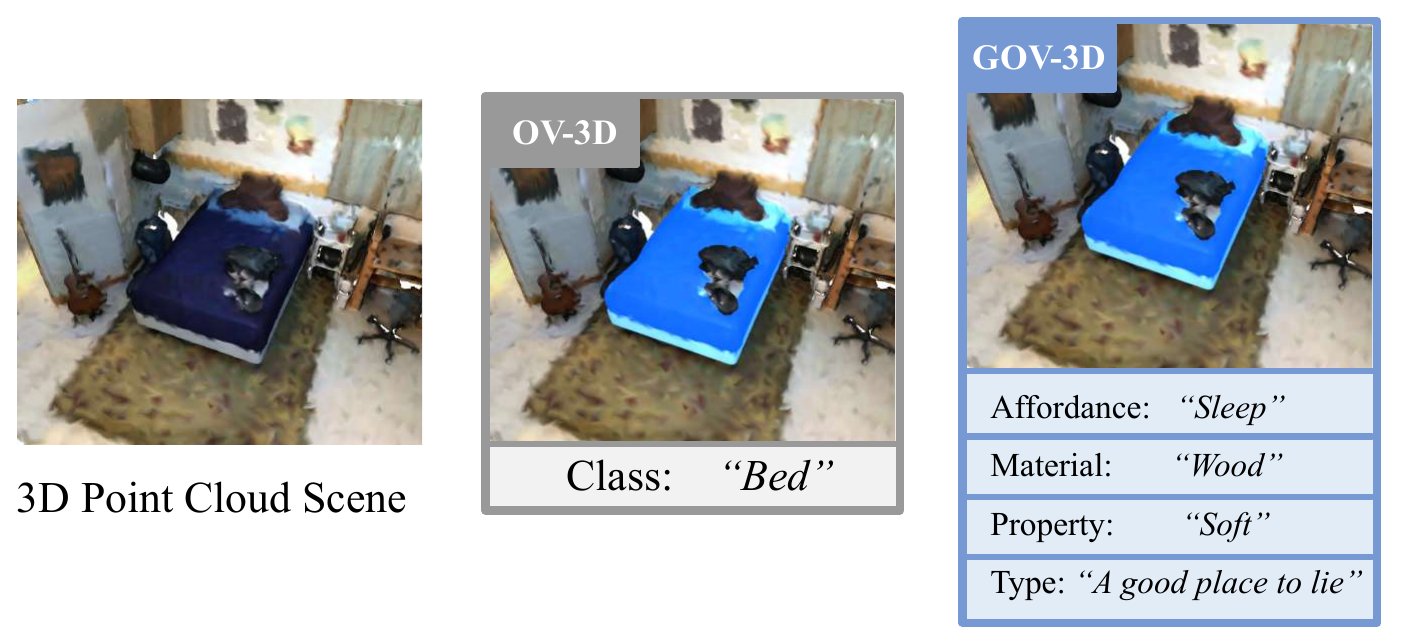}
        \vspace{-2mm}
            \caption{\zyj{The proposed Generalized Open-Vocabulary 3D Scene Understanding (GOV-3D) \ryn{task} expands the vocabulary types of the classic 3D Scene Understanding (OV-3D) task. \ryn{While} OV-3D only}  
         \ryn{supports queries of object classes}, \zyj{GOV-3D} supports queries of %
         object-related abstract attributes.} %
            \label{teasor_compar}
            \vspace{-4mm}
\end{figure}

\rl{In this paper, we aim to study how well current OV-3D methods can generalize their understanding beyond 3D object classes to  open-set object attribute vocabularies.} Specifically, we introduce a more challenging task called Generalized Open-Vocabulary 3D Scene Understanding (GOV-3D). GOV-3D takes a 3D point cloud scene and a text query as input to predict a corresponding 3D mask of the best matching object, which is the same as OV-3D. \ryn{However, unlike OV-3D which only supports object classes as the input} text query, GOV-3D supports abstract vocabularies that specify the attribute of the target object in the \ryn{input} text query, as shown in \zyj{Figure~\ref{teasor_compar}.} This requires a comprehensive understanding of both \ryn{3D objects} and 3D scenes, making the \ryn{GOV-3D} task more challenging in practical scenarios.

Existing 3D scene understanding benchmarks, \zyj{such as ScanNet~\cite{scannet}, ScanNet200~\cite{scannet200}, and ScanNet++~\cite{scannet++}}, only provide annotations for object classes.
To address \ryn{this limitation of existing benchmarks}, we construct a new benchmark, named \textit{OpenScan}, for the GOV-3D task. OpenScan is constructed based on the ScanNet200~\cite{scannet200} benchmark. It expands the \ryn{single category of object classes in ScanNet200 into \zyj{eight linguistic aspects}} of object-related attributes, including \ryn{\textit{affordance}, \textit{property}, \textit{\jy{type}}, \textit{manner}, \textit{synonym}, \textit{requirement}, \textit{element}, and \textit{material}}.
\ysq{\ryn{This allows each object to be associated with some} generalized knowledge beyond object \ryn{classes}.}
With our OpenScan benchmark, it becomes possible to comprehensively evaluate existing OV-3D models from various aspects, enabling a quantitative assessment of their generalization capabilities in understanding abstract object attributes. 

We \ryn{have} compared \zyj{seven} strong baseline methods \ryn{under the GOV-3D task, on our} OpenScan benchmark. Experimental results demonstrate that the current state-of-the-art OV-3D models excel in understanding basic object classes, but significantly \ryn{degrade in their} ability to understand object attributes, such as \zyj{affordance and material.} 
This highlights the importance of establishing a comprehensive and reliable benchmark to identify the weaknesses of OV-3D models. %
\ryn{The key contributions of this work can be summarized as}:
\begin{itemize}[leftmargin=*]
\item We introduce a challenging task of Generalized Open-Vocabulary 3D Scene Understanding (GOV-3D) \zyj{to extend the classic OV-3D task for a more general understanding of 3D scenes.}
\item We provide a novel benchmark named OpenScan for the GOV-3D task, which \ryn{facilitates comprehensive evaluation of} the generalization ability of \zyj{OV-3D} segmentation models on abstract object attributes.
\item We conduct extensive experiments \ryn{with} existing \zyj{OV-3D} segmentation models \ryn{on} our OpenScan benchmark, showing that \zyj{even the latest \ryn{methods}} struggle to understand the abstract object attributes beyond object classes.
\end{itemize}

\section{Related Work}

\noindent\textbf{Open-Vocabulary 3D Scene Understanding.}
The study of open-vocabulary 3D scene understanding~\cite{hcma} has been relatively limited compared to open-vocabulary 2D understanding. This is primarily due to the complexity and difficulty in obtaining 3D datasets. OpenMask3D~\cite{takmaz2023openmask3d} introduces the open-vocabulary 3D instance segmentation task. It propose\jy{s} the first approach for \rl{the task} in a zero-shot setting. OpenScene~\cite{peng2023openscene} also proposes a zero-shot method for open-vocabulary 3D scene understanding. Beyond the object class, it is able to utilize arbitrary text queries for semantic segmentation. Previous methods have mainly focused on object context for 3D scene understanding. PLA~\cite{ding2023pla} and RegionPLC~\cite{regionplc} extend the context to a more coarse-to-fine semantic representation to provide a more comprehensive supervision. Recently, OpenIns3D~\cite{huang2024openins3d}, Open3DIS~\cite{nguyen2023open3dis}, and SAI3D~\cite{sai3d} utilize powerful 2D segmentation models to generate 2D instances and then merge them into 3D instances. Instead of utilizing accurate 2D masks from 2D segmentation models, MaskClustering~\cite{yan2024maskclustering} leverages clustering algorithms to perform zero-shot 3D segmentation. \zyjn{Recently, UniSeg3D~\cite{UniSeg3D} proposes a unified framework for 3D scene understanding.} However, these methods only provide qualitative results for object attributes and lack a thorough evaluation of performance beyond object classes. This motivates us to conduct a quantitative evaluation that encompasses a wider range of object attributes.

\noindent\textbf{3D Scene Understanding Benchmark.}
\zyjn{\ryn{Existing open-vocabulary 3D scene understanding benchmarks, \rl{\eg,} ScanNet~\cite{scannet}, ScanNet200~\cite{scannet200}, S3DIS~\cite{s3dis}, and Matterport3D~\cite{Matterport3D}, utilize RGB-D cameras, while ARKitScenes~\cite{dehghan2021arkitscenes}, Replica~\cite{replica}, and ScanNet++~\cite{scannet++} leverage high-resolution laser scanners to capture high-fidelity 3D data for 3D reconstructions}. \zyja{Our proposed OpenScan benchmark expands the object class annotations of the open-vocabulary 3D scene understanding benchmarks (\ie, ScanNet200) to object-related attributes.}
\zyja{Similar to the 3D referring~\cite{chen2020scanrefer} and 3D reasoning~\cite{huang2025reason3d} benchmarks, our OpenScan introduces new annotations for existing 3D scans to locate 3D objects via text queries. Unlike these tasks, which assume the queried object exists in the scene, our introduced GOV-3D task requires discriminative capabilities to determine whether the query presents in the scene.}
MMScan~\cite{lyu2024mmscan} provides a benchmark for visual attribute understanding but lacks commonsense-related attribute annotations (\eg, ``\textit{synonym}'' and ``\textit{requirement}'') included in our OpenScan.
Recently, SceneFun3D~\cite{delitzas2024scenefun3d} provides a large-scale 3D dataset with annotations for functionality and affordance interactions in 3D scenes. However, our OpenScan differs from SceneFun3D by considering a broader range of attributes. Specifically, while SceneFun3D focuses on function or affordance understanding, our OpenScan covers eight linguistic aspects, with affordance representing just one of them. Besides, while SceneFun3D focuses on element-level human-scene interaction (\eg, ``\textit{door handle}''), our OpenScan is designed for object-level scene understanding (\eg, ``\textit{door}'').
}

\begin{figure*}[h] \centering
    \includegraphics[width=1\textwidth]{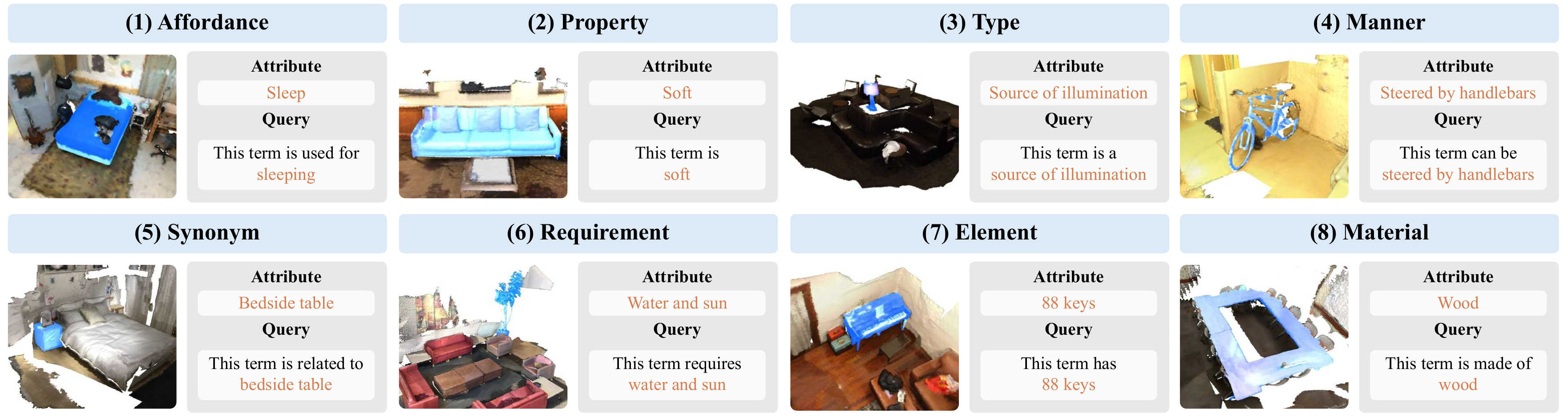}
    \vspace{-5mm}
    \caption{OpenScan benchmark samples. The target \rl{objects are} highlighted in blue.}
    \label{dataset_vis}
    \vspace{-5mm}
\end{figure*}

\section{Task Setting and Benchmark}

\subsection{Task Formulation}

\noindent\textbf{OV-3D.}
Let \zyj{$P=\{p_n\}_{n=1}^{N} \in \mathbb{R}^{N \times 3}$} \jy{represent \zyj{3D scenes with $N$ points}\zyja{, $I=\{i_{x}\}_{x=1}^{X} \in \mathbb{R}^{H \times W \times 3}$ denote $X$ RGB image frames,} and $V=\{c_{t}\}_{t=1}^{T}$ is a vocabulary set of $T$ text sentences, each describing \zyj{the object class \zyj{$c_t$}} \rl{that} we aim to detect.} An OV-3D model, \zyj{$\mathbb{M}$}, generates predictions \rl{with high confidence scores,} \zyja{$Q = \zyj{\mathbb{M}}(P, I, V)$}. \zyj{\rl{Predictions $Q$ are then compared with the GT} label $G$ for evaluation.}

\noindent\textbf{GOV-3D.}
\ysq{The existing 3D scene understanding benchmark, denoted as $\mathcal{D} = \{(o_k, c_k)\}_{k=1}^{K}$, comprises a collection of $K$ object-label pairs. Each pair consists of an object $o_k$ represented as a \zyja{3D mask} and its corresponding class label $c_k$. The benchmark is composed of multiple 3D scenes \zyja{$P=\{p_n\}_{n=1}^{N} \in \mathbb{R}^{N \times 3}$ with $N$ points, and $X$ RGB image frames $I=\{i_{x}\}_{x=1}^{X} \in \mathbb{R}^{H \times W \times 3}$.}
Building upon this, GOV-3D extends the class label \zyj{$c_{k}$} to object attribute \zyj{$a_{k}$}. Accordingly, the attribute set for 3D scenes $P$ is defined as \rl{$A=\{a_k\}_{k=1}^{H}$, composed of $H$ text sentences, each describing 
a specific attribute $a_k$ that} we aim to detect.}
\jy{A GOV-3D model, $\mathbb{N}$, produces predictions \rl{with high confidence scores,} \zyja{$Q=\mathbb{N}(P,I,A)$}}. \zyj{The evaluation of the GOV-3D task involves comparing the predictions $Q$ \rl{with} the GT label $G$.}

\subsection{\zyj{Benchmark Description}}

The OpenScan benchmark is constructed based on the ScanNet200~\cite{scannet200} benchmark, \zyj{which consists of 200 object classes with more than 1,500 3D scans. Since the ScanNet200 benchmark \rl{contains only an} \jy{object-level} class annotation for each object, it is \rl{not suitable for our} GOV-3D task. To perform the GOV-3D task, we construct the OpenScan benchmark by leveraging the object \rl{annotations} of the ScanNet benchmark.} \jy{Our OpenScan provides} attribute \jy{annotations} for each object, expanding the single category of object classes in ScanNet200 into eight linguistic aspects of object-related attributes, including \textit{affordance}, \textit{property}, \textit{type}, \textit{manner}, \textit{synonym}, \textit{requirement}, \textit{element}, and \textit{material}. Figure~\ref{dataset_vis} shows an example from our OpenScan benchmark. \jy{The target objects in our OpenScan are} annotated with eight linguistic aspects of object attributes.
\jy{The \rl{explanations} of these eight object attributes are as follows}:
\begin{itemize}[leftmargin=*]
\item \textbf{\textit{Affordance}}: \rl{\zyjn{is} the object function or usage (\eg, \textit{``sit''} for a chair)}.
\item \textbf{\textit{Property}}: \rl{indicates the object characteristic (\eg, \textit{``soft''} for a \zyjn{pillow})}.
\item \textbf{\textit{Type}}: \rl{indicates the object category or group (\eg, \textit{``a communication device''} for a telephone)}.
\item \textbf{\textit{Manner}}: \rl{indicates the object behavior \zyjn{(\eg, \textit{``worn on a head''} for a hat)}}.
\item \textbf{\textit{Synonym}}: \zyjn{\rl{is a term with a} similar meaning (\eg, \textit{``image''} for a picture)}.

\item \textbf{\textit{Requirement}}: indicates %
\rl{an essential condition} that an object should possess to fulfill a specific need \zyjn{(\eg, ``\textit{balance to ride}'' \rl{for a bicycle})}. 
\item \textbf{\textit{Element}}: \ryn{indicates an individual component or part that constitutes the object \zyjn{(\eg, ``\textit{two wheels}'' \rl{for a bicycle})}.}
\item \textbf{\textit{Material}}: \ryn{indicates the type of material of the object \zyjn{(\eg, ``\textit{plastic}'' for a bottle)}}.
\end{itemize}

\subsection{Benchmark Annotation}

\begin{figure*}[t] \centering
    \includegraphics[width=0.95\textwidth]{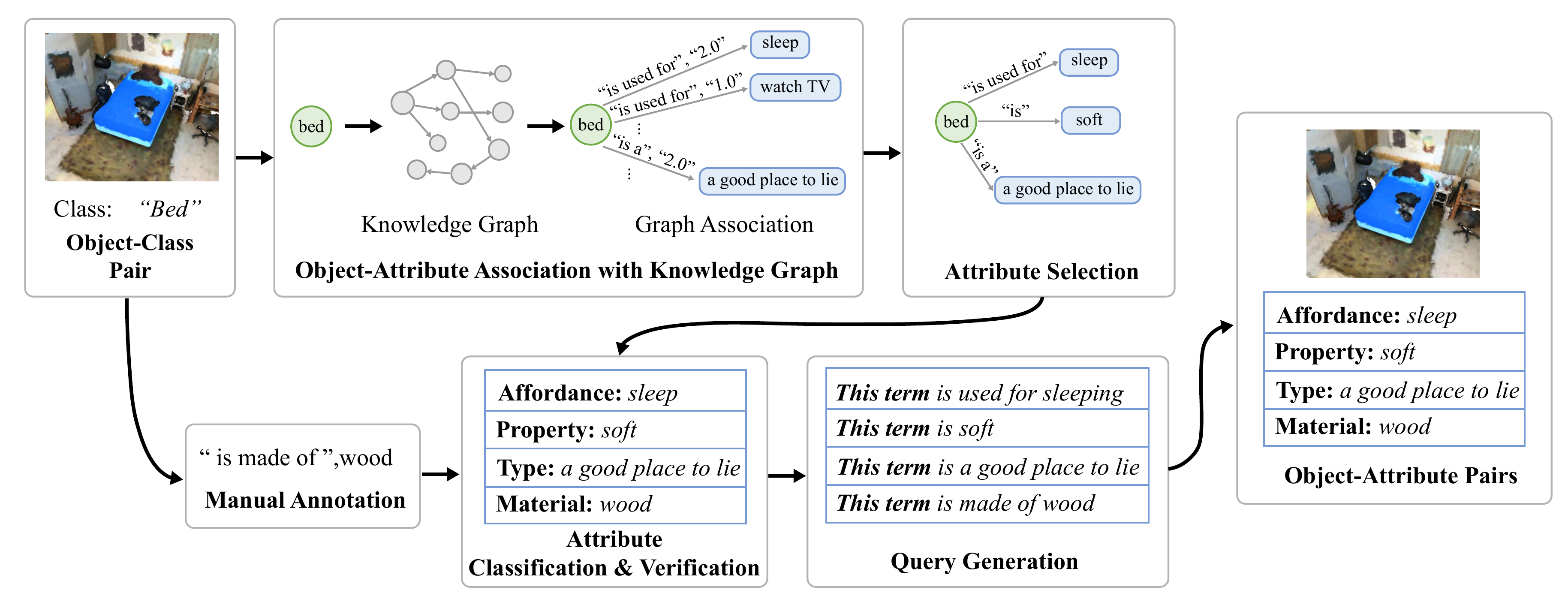}
    \vspace{-4mm}
    \caption{Illustration of \jy{the} data generation process \rl{for} our OpenScan benchmark.}
    \label{dataset_generation}
    \vspace{-5mm}
\end{figure*}

\jy{Figure~\ref{dataset_generation} illustrates the annotation process of our OpenScan benchmark.} We first leverage the knowledge graph to establish \jy{the} association between objects and various attributes. \jy{We also conduct manual annotations to label the visual \rl{attributes} of each object}. \zyjn{Finally, we classify and verify these attributes to ensure semantic consistency.}

\noindent\textbf{Object-Attribute Association with Knowledge Graph.}
\ysq{We associate each object with various attributes using knowledge graphs, as illustrated in} Figure~\ref{dataset_generation}. \ysq{Let \zyj{$\mathcal{D}= \{(p_{k},c_{k})\}_{k=1}^{K}$} \rl{denotes} the existing 3D scene understanding benchmark, \eg, ScanNet200~\cite{scannet200} in our implementation, where \zyj{$p_{k}$} is a target object, \zyj{$c_{k}$} is the corresponding class label, and \zyj{$K$ denotes the number of target object and label pairs.}} \zyj{The benchmark is composed of multiple 3D scenes \zyj{$P \in \mathbb{R}^{N \times 3}$ with} \zyj{$N$ points}}. \ysq{Let $\mathcal{G}=(\mathcal{V}, \mathcal{E})$ \rl{denotes the knowledge graph, where $\mathcal{V}$ is the node set and $\mathcal{E}$} is the edge set}. \ysq{The nodes $v\in\mathcal{V}$ are natural language words and phrases, and the edges $e\in\mathcal{E}$ are relation knowledge connecting them. Each edge $e$ is directional, and can be represented as a tuple $(v_{m}, \zyj{r, w,} v_{n})$, where $v_{m}, v_{n} \in \mathcal{V}$ are the \rl{names} of the head node and the tail node, \zyj{$r$} is the relation, and \zyj{$w$} is the importance weight of this relation.} \ysq{We extract the relation knowledge from the \rl{popular and high-quality} NLP knowledge base ConceptNet~\cite{speer2017conceptnet}. An example of relation knowledge from it is:}
\begin{equation}
\ysq{e = (\text{``\textit{bed}''}, \text{``\textit{is used for}''}, 2.0, \text{``\textit{sleep}''}).}
\end{equation}
\ysq{We query a set of relation knowledge $\{e\}_{i}$ linked to object class $c_{i}$ from the knowledge graph $\mathcal{G}$. Formally, for each edge within it, the head node name is the same as the input object class, \ie, $v_{m}=c_{i}$. The query process is defined as:}
\begin{equation}
\ysq{\{e\}_{i} = \left\{(v_{m}, r, w, v_{n})\in \mathcal{E} | v_{m}=c_{i} \right\}.}
\end{equation}

\noindent\textbf{Attribute Selection.}
\zyj{In the set of relation knowledge $\mathcal{E}$, \ysq{we keep the attribute with the highest weight $w$ in the same relation \zyj{$r$}}. Given a relation knowledge $e_{i} \in \{e\}$, we have:
\begin{equation}
\{e\}^{'}_{i}=\{e_{i}|r_{j} = r_{i} \wedge \forall e_{j} \in \{e\}: w_{j} \leq w_{i}\}.
\end{equation}
These object-attribute pairs \ysqq{form the basic annotations of OpenScan}, which is useful in the GOV-3D task. Finally, each 3D object $p_{i}$ is assigned a relation knowledge $e_{i}$ through $I$ annotations, serving as commonsense knowledge \rl{$\mathcal{Y}_{c}$ as}:} 
\begin{equation}
\mathcal{Y}_{c}= \{(p_{i}, e_{i}) ,e_{i}\in \{e\}^{'} | v_{m}=c_{i}\}_{i}^{I}.
\end{equation}

\noindent\textbf{Manual Annotation.}
For the visual attribute \jy{that cannot be inferred without human perception,} \ysqq{we manually annotate each 3D object} \jy{following a rigorous protocol}. \zyjn{We create a web interface for annotators to select \ysqq{each object's visual attribute}.} Specifically, for each scene, annotators \ysqq{view the 3D point cloud and the target’s corresponding 2D image}. \jy{Taking the \textit{material} attribute as an example, annotators are tasked with identifying the primary material composition of the target object.}
Any 3D object \jy{with an} ambiguous appearance \rl{is} carefully identified through \jy{different camera views} of the scene and \jy{the corresponding image frames around the object}. \jy{Finally, each 3D object \zyj{$p_{j}$} is assigned with a relation \zyj{$r_{j}=``is\ made\ of"$} and a visual attribute like \rl{\textit{material}} \zyj{$v_{n}$ through $J$ annotations}, serving as visual appearance $\mathcal{Y}_{m}$} \jy{in:}
\begin{equation}
\mathcal{Y}_{m}=\{(p_{j},(r_{j}, v_{n})\}_{j}^{J}.
\end{equation}
\jy{After obtaining the attribute annotations based on commonsense knowledge $\mathcal{Y}_{c}$ and visual appearance $\mathcal{Y}_{m}$ of the 3D objects}, we \jy{use the combination of} these two categories of attributes \jy{as the whole annotations $\mathcal{Y}$ for our OpenScan}.

\noindent\textbf{Attribute Classification.}
To better organize our benchmark, we \zyj{manually} \ysqq{group each attribute into eight linguistic aspects based on} relation \zyj{$r$} 
and attribute $v_{n}$. \zyj{This process involves considering the nature of the relation \zyj{$r$} and attribute $v_{n}$, and how they align with each linguistic aspect.} 
\zyj{Subsequently}, each attribute $v_{n}$ is assigned to a linguistic aspect.

\noindent\textbf{Attribute Verification.}
After the initial attribute classification, we manually verify each 3D object $p_{k}$ with its corresponding attribute $v_{n}$ and linguistic aspect. If a 3D object $p_{k}$ contains multiple attributes $v_{n}$ within a single linguistic aspect, we manually assign the most representative attribute to ensure evaluation consistency. This attribute is uniquely tied to the object class of $p_{k}$ to eliminate cross-class ambiguity.
\rl{We also} filter out similar attributes $v_{n}$ (\eg, ``store things'' and ``store somethings''), preserving only one attribute to ensure consistency. \zyja{During the verification process, the attributes across eight linguistic aspects are reduced from 528 to 341.}

\noindent\textbf{Query Generation.}
\jy{A} practical \ysqq{GOV-3D query} should \jy{incorporate} attribute names but \jy{exclude object names}\ysqq{, requiring a query strategy that} focuses on attributes rather than exposing object identities (\textit{i.e.}, object classes). To achieve this, we perform query generation by hiding the object \zyj{class $v_{m}$} of the object \zyj{$p_{k}$}. We first replace the object \zyj{classes $v_{m}$} with a substitution term $t=\text{``\textit{this term}''}$. \jy{\rl{The substitution term $t$, \zyj{the relation $r$, and the corresponding attribute $v_{n}$} are then} concatenated to form the text query \zyj{$q$} as}:
\begin{equation}
q = Concate(t, r, v_{n}).
\end{equation}
In this way, we generate text queries \zyj{$q$} that correspond to object-attribute \zyj{annotations
$\mathcal{Y}$}.
We then perform manual verification again on text queries. With text queries as input, we \rl{can} conduct evaluations on existing OV-3D models.

\begin{table*}[t]
    \centering
    \resizebox{1\textwidth}{!}
    {
    \begin{tabular}{l|cccccccc|c}
        \toprule
\textbf{Statistics}	& \textbf{Affordance} & \textbf{Property}  &	\textbf{Type} &	\textbf{Manner} &	\textbf{Synonym}  &  \textbf{Requirement} &	\textbf{Element}	& \textbf{Material} &	\textbf{All} \\
\midrule
Attributes	& 104 & 19 & 96 & 21 & 16 & 28 & 47 & 10 & 341
 \\

\zyjn{Attribute} Annotations &	37,362 & 8,591 &  28,293 & 4,925 & 2,937 & 9,695 & 13,505 & 48,336 & 153,644
 \\
Attribute Annotations per Object &	0.77 & 0.18 &  0.58 & 0.10 & 0.06 & 0.20 & 0.28 & 0.99 & 3.15
 \\
Attribute Annotations per Scene &	24.69 & 5.68 &  18.70 & 3.26 & 1.94 & 6.41 & 8.93 & 31.95 & 101.55
 \\
        \bottomrule
    \end{tabular}
    }
    \vspace{-3mm}
    \caption{OpenScan benchmark statistics of object-related attributes for \rl{the} eight linguistic aspects.}
\label{dataset_statistics}
\vspace{-4mm}
\end{table*}

\subsection{Benchmark Statistics}
\rl{Table~\ref{dataset_statistics} shows the statistics of our OpenScan benchmark. We have} collected eight linguistic aspects of attributes, providing a total of \zyj{153,644} attribute annotations across 341 attributes for 1,513 scenes \zyjn{in ScanNet200~\cite{scannet200}. In these aspects, the visual aspect of \textit{material} is annotated manually, while other attributes are automatically generated via knowledge graph. There are \zyj{101.55} attribute \zyjn{annotations} per scene in 1,513 3D indoor scenes. Besides, each object is annotated with an average of 3.15 attributes, indicating that most objects in ScanNet200 receive multiple attribute labels and are comprehensively represented.} While certain \zyj{linguistic aspects} such as \textit{manner} and \zyjn{\textit{synonym}} encompass a limited number of attributes, others like \textit{affordance} and \textit{type} consist of a wide range of attributes. We follow the training and validation split settings of ScanNet200.

\subsection{Evaluation Metrics}
We employ commonly used OV-3D metrics to evaluate our GOV-3D task. For semantic segmentation, we follow~\cite{peng2023openscene, ding2023pla, regionplc} to apply mean IoU (mIoU) and mean accuracy (mAcc). For instance segmentation, we follow~\cite{takmaz2023openmask3d, sai3d, yan2024maskclustering, nguyen2023open3dis} to apply average \zyja{precision} (AP) at IoU scores of 25\% (AP$_{25}$), 50\% (AP$_{50}$), and the mean of AP from 50\% to 95\% at 5\% steps.
\zyja{We compute the mean score (Mean) across all attributes to obtain the overall performance.}
\section{Experiments}

\begin{table}[t]
\centering
    \resizebox{0.46\textwidth}{!}
    {
    \begin{tabular}{l|ccc}
        \toprule
        
      \multirow{2}{*}{\textbf{Model}}  & \multirow{2}{*}{\textbf{Training}} & \textbf{Pre-Trained} & \textbf{Pre-Trained} \\
      & &  \textbf{3D Proposal} & \textbf{2D Proposal} \\
        \midrule
        OpenMask3D & - & Mask3D & SAM \\
        SAI3D & - & - & Semantic-SAM \\
        MaskClustering & - & - & CropFormer  \\
        Open3DIS & - & ISBNet & Grounded-SAM \\
        OpenScene & - & - & -\\
        PLA & ScanNet & - & -\\
        RegionPLC & ScanNet & - & -\\
        
        \bottomrule
    \end{tabular}
    }
    \vspace{-3mm}
    \caption{The detailed information of the OV-3D models.}
\label{baseline_setting}
\vspace{-4mm}
\end{table}

We conduct \rl{our experiments} on the validation set of our OpenScan across \jy{eight} \zyj{linguistic aspects} using the publicly available OV-3D models. \zyja{All OV-3D models are evaluated in a zero-shot setting without training on the OpenScan benchmark.} For 3D instance segmentation, we evaluate OpenMask3D~\cite{takmaz2023openmask3d}, SAI3D~\cite{sai3d}, MaskClustering~\cite{yan2024maskclustering}, and Open3DIS~\cite{nguyen2023open3dis}. For 3D semantic segmentation, we evaluate OpenScene~\cite{peng2023openscene}, PLA~\cite{ding2023pla}, and RegionPLC~\cite{regionplc}. \rl{Table~\ref{baseline_setting} summarizes the information of these models: \jy{the} training set, \zyj{\zyjn{pre-trained} 3D proposal, and \zyjn{pre-trained} 2D proposal}.} \zyjn{All experiments are conducted on one NVIDIA RTX 4090 GPU.}

\subsection{Main Results}
\noindent\textbf{3D Instance Segmentation.}
We evaluate OpenMask3D~\cite{takmaz2023openmask3d}, SAI3D~\cite{sai3d}, MaskClustering~\cite{yan2024maskclustering}, and Open3DIS~\cite{nguyen2023open3dis} across \jy{341 attributes from our OpenScan and 198 object class\jy{es} from ScanNet200~\cite{scannet200}}. 
\zyjn{Table~\ref{main_result} show that the \rl{performance of these OV-3D models on our GOV-3D benchmark, OpenScan, are significantly lower than those} on the classic OV-3D dataset, ScanNet200. This gap underscores that our proposed GOV-3D task is a more challenging extension of the traditional OV-3D task.}

When comparing the results of each OV-3D model across different linguistic aspects, we observe \rl{higher performances} in the \textit{synonym} and \textit{material} aspects but struggle in the \textit{affordance} and \textit{property} aspects. \zyj{The high performance in the \textit{synonym} aspect can be attributed to the close similarity between attributes in this aspect and object classes, making recognition easier compared to the more abstract \textit{affordance} and \textit{property} aspects. An example of these closely related terms is shown in Figure~\ref{dataset_vis}, where the corresponding \textit{synonym} aspect of the object class ``\textit{nightstand}'' is ``\textit{bedside table}''.
The high performance in the \textit{material} aspect highlights the ability of these OV-3D models to recognize visual patterns. By utilizing CLIP~\cite{clip} for 3D scene understanding, these models benefit from its visual patterns, including material and color from 
\ysqq{image-text pretraining}
, enhancing their comprehension of visual attributes beyond other attributes.}

When comparing the results of each linguistic aspect in our OpenScan to \rl{those of the} object class in ScanNet200, we notice that certain aspects like \textit{synonym} and \textit{material} perform even better than the object class. This can be attributed to the smaller number of attributes in these two aspects when compared to the broader and more diverse set of object classes. 
A smaller set of classes can increase the model's confidence in its predictions, facilitating more accurate predictions without the complexity of distinguishing \rl{among} a large number of categories.
Notably, Open3DIS shows impressive results in various linguistic aspects compared to other OV-3D models, aligning with its strong \rl{performances} \jy{in \rl{the classic OV-3D task} (\ie, evaluating only \rl{on} object classes)}.

\begin{table*}[t]
\renewcommand\arraystretch{0.85}
\fontsize{16pt}{26pt}\selectfont

    \resizebox{0.98\textwidth}{!}
    {
    \begin{tabular}{l|cccccccc|c|c}
        \Xhline{1.4pt}
      \multirow{2}{*}{\textbf{Method}} & \multicolumn{9}{c|}{\textbf{OpenScan}} & \textbf{ScanNet200} \\
        \Xcline{2-11}{0.5pt}
      & \textbf{Affordance} &  \textbf{Property} & \textbf{Type} & \textbf{Manner} & \textbf{Synonym} & \textbf{Requirement} & \textbf{Element} & \textbf{Material} & \textbf{Mean} & \textbf{Object Class} \\
        \Xhline{1pt}
        \multicolumn{11}{c}{\textbf{AP}} \\
        \Xhline{1pt}
        OpenMask3D~\cite{takmaz2023openmask3d} & 7.2 & 7.5 & 8.5 & 12.8 & 16.9 & 13.0 & 12.2 & 18.8 & 9.9 & 15.4\\
        SAI3D~\cite{sai3d} & 5.3 & 5.8 & 7.8 & 11.3 & 10.0 & 10.0 & 8.7 & 11.3 & 7.7 & 12.7\\
        MaskClustering~\cite{yan2024maskclustering} & 6.2 & 7.0 & 7.1 & 11.1 & 16.2 & 11.3 & 7.4 & 12.1 & 8.1 & 12.0\\
        Open3DIS~\cite{nguyen2023open3dis} & \textbf{11.9} & \textbf{12.8} & \textbf{14.2} & \textbf{19.2} & \textbf{26.7} & \textbf{19.2} & \textbf{18.7} & \textbf{28.3} & \textbf{15.8} & \textbf{23.7}\\
        \Xhline{1pt}
        \multicolumn{11}{c}{\textbf{AP$_{50}$}} \\
        \Xhline{1pt}
        OpenMask3D~\cite{takmaz2023openmask3d} & 9.1 & 10.0 & 11.2 & 15.4 & 19.7 & 16.0 & 15.4 & 22.1 & 12.5 & 19.9\\
        SAI3D~\cite{sai3d} & 8.4 & 8.3 & 11.4 & 15.7 & 16.7 & 15.3 & 13.6 & 17.1 & 11.6 & 18.8\\
        MaskClustering~\cite{yan2024maskclustering} & 10.7 & 12.3 & 13.3 & 18.4 & 30.3 & 21.8 & 13.5 & 20.6 & 14.6 & 23.3\\
        Open3DIS~\cite{nguyen2023open3dis} & \textbf{14.8} & \textbf{16.0} & \textbf{17.9} & \textbf{22.3} & \textbf{30.6} & \textbf{24.1} & \textbf{21.9} & \textbf{33.6} & \textbf{19.3} & \textbf{29.4}\\
        \Xhline{1pt}
        \multicolumn{11}{c}{\textbf{AP$_{25}$}} \\
        \Xhline{1pt}
        OpenMask3D~\cite{takmaz2023openmask3d} & 10.4 & 11.6 & 13.0 & 17.4 & 20.6 & 18.9 & 17.1 & 25.0 & 14.2 & 23.1\\
        SAI3D~\cite{sai3d}  & 10.5 & 10.7 & 13.4 & 18.2 & 20.0 & 18.7 & 16.0 & 22.9 & 14.1 & 24.1\\
        MaskClustering~\cite{yan2024maskclustering} & 13.7 & 15.8 & 17.7 & 23.1 & \textbf{36.6} & \textbf{28.2} & 17.2 & 25.6 & 18.7 & 30.1\\
        Open3DIS~\cite{nguyen2023open3dis} & \textbf{16.7} & \textbf{16.8} & \textbf{20.2} & \textbf{24.2} & 33.1 & 25.5 & \textbf{24.7} & \textbf{36.7} & \textbf{21.4} & \textbf{32.8}\\
        \Xhline{1.5pt}
    \end{tabular}
    }
    \vspace{-2mm}
    \caption{3D instance segmentation results on our OpenScan benchmark.} 

    \vspace{-3mm}
\label{main_result}
\end{table*}

\begin{table}[t]
\small
\centering
\renewcommand\arraystretch{1.2}
    \resizebox{0.45\textwidth}{!}
    {
    \begin{tabular}{l|cc|cc}
        \toprule
        \multirow{2}{*}{\textbf{Method}} & \multicolumn{2}{c|}{\textbf{OpenScan}} & \multicolumn{2}{c}{\textbf{ScanNet}} \\
        & \textbf{mIoU} & \textbf{mAcc} & \textbf{mIoU} & \textbf{mAcc} \\
        \midrule
        OpenScene~\cite{peng2023openscene} & \textbf{0.45} & 1.87 & 47.5 & 70.7 \\
        PLA~\cite{ding2023pla} & 0.01 & \textbf{2.37} & 66.6 & 77.5 \\
        RegionPLC~\cite{regionplc} & 0.07 & 2.36 & \textbf{68.7} & \textbf{78.7} \\
        \toprule
    \end{tabular}
    }
    \vspace{-2mm}
\caption{3D semantic segmentation results on OpenScan.
}
\label{semantic_result}
\vspace{-3mm}
\end{table}

\noindent\textbf{3D Semantic Segmentation.}
We evaluate OpenScene~\cite{peng2023openscene}, PLA~\cite{ding2023pla}, and RegionPLC~\cite{regionplc}, \rl{reporting the average score of} all attributes \zyj{in our OpenScan and that of all object classes in ScanNet~\cite{scannet}}. Table~\ref{semantic_result} shows that although these OV-3D models \jy{perform well in} recognizing object classes, they exhibit poor \rl{performances} on linguistic aspects with low mIoU and mAcc metrics.
\ysq{The methods for semantic segmentation suffer \jy{from} a more significant performance drop on OpenScan \jy{when} compared \jy{with} those for instance segmentation. \zyj{This drop can be \jy{caused by} several factors. \rl{First, there is a significant discrepancy in the vocabulary size} between ScanNet and our OpenScan. 
A larger vocabulary size implies a more diverse set of semantic concepts that the model needs to comprehend, making our OpenScan more challenging \jy{and practical in real-world scenarios}. \rl{In addition}, the arbitrary nature of object attributes in contrast to object classes adds complexity to the GOV-3D task. Besides, the lack of both robust 3D proposals (\eg, Mask3D~\cite{mask3d}) and 2D proposals (\eg, SAM~\cite{sam}) for class-agnostic masks} can also \jy{be attributed} to the drop.}
\zyj{Conversely, instance segmentation models like OpenMask3D~\cite{takmaz2023openmask3d} leverage strong \jy{instance-level knowledge, \eg, proposals extracted from Mask3D and SAM}, to effectively segment novel 3D objects, leading to higher \rl{performances} on the GOV-3D task.}

\begin{figure}[tb] \centering
    \includegraphics[width=0.46\textwidth]{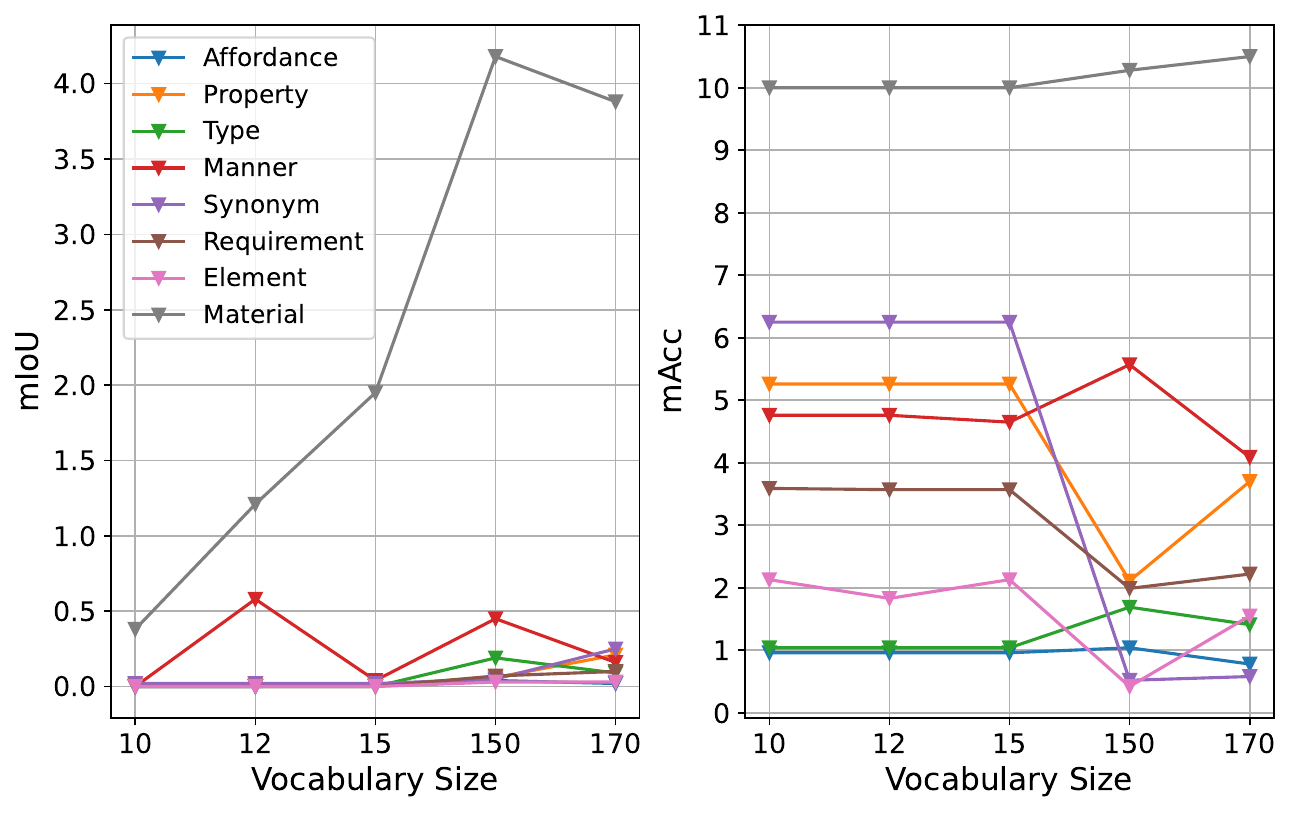}
    \vspace{-3mm}
    \caption{\rl{Impact of different pre-training vocabulary size}.}
    \label{vocabulary_size}
    \vspace{-5mm}
\end{figure}

\begin{table}[t]
\small
\centering
\renewcommand\arraystretch{1.2}
    \resizebox{0.45\textwidth}{!}
    {
    \begin{tabular}{l|c|ccc}
        \toprule
        \textbf{Method} & \textbf{Template} & \textbf{AP} & \textbf{AP$_{50}$} & \textbf{AP$_{25}$} \\
        \midrule
        \multirow{2}{*}{OpenMask3D~\cite{takmaz2023openmask3d}} & - & 9.7 & 12.2 & 14.1 \\
        & \checkmark & \textbf{9.9} & \textbf{12.5} & \textbf{14.2} \\
        \midrule
        \multirow{2}{*}{SAI3D~\cite{sai3d}} & - & 6.7 & 10.1 & 12.8 \\ 
        & \checkmark & \textbf{7.7} & \textbf{11.6} & \textbf{14.1}\\
        \midrule
        \multirow{2}{*}{MaskClustering~\cite{yan2024maskclustering}} & - & 6.8 & 12.0 & 14.6 \\   
        & \checkmark & \textbf{8.1} & \textbf{14.6} & \textbf{18.7}\\
        \midrule
        \multirow{2}{*}{Open3DIS~\cite{nguyen2023open3dis}} & - & 15.6 & 19.2 & 21.3 \\   
        & \checkmark & \textbf{15.8} & \textbf{19.3} & \textbf{21.4}\\
        \toprule
    \end{tabular}
    }
\vspace{-2mm}
\caption{Effects of query form on our OpenScan benchmark.}
\label{query_form}
\vspace{-4mm}
\end{table}

\begin{figure*}[tb] \centering
    \includegraphics[width=0.9\textwidth]{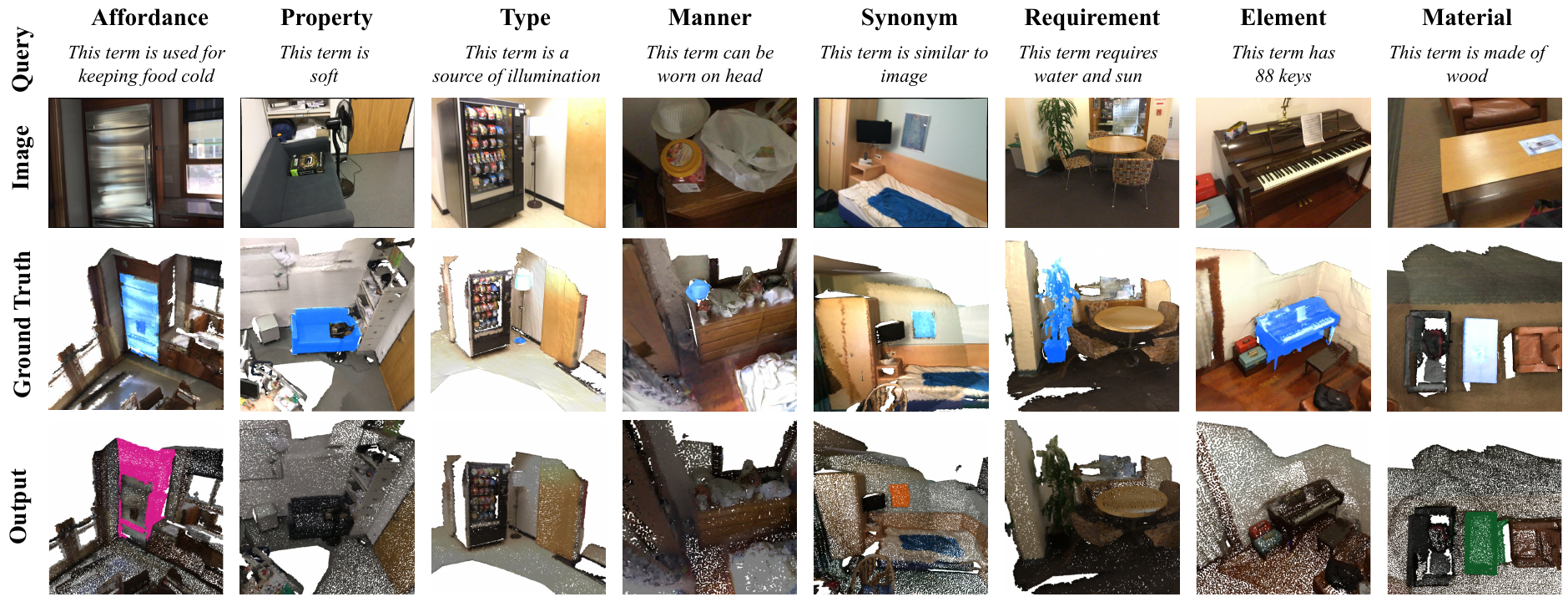}
    \vspace{-3mm}
    \caption{\zyjn{Qualitative results of Open3DIS \rl{on} our OpenScan benchmark. The GT objects and outputs are highlighted in color.}}
    \label{qualitative_supp}
    \vspace{-4mm}
\end{figure*}

\subsection{The Impact of the Pre-trained Vocabulary Size}

\rl{To study the impact of the pre-trained vocabulary size \zyja{(\ie, the number of pre-trained object classes)} on the GOV-3D task, we conduct experiments based on RegionPLC~\cite{regionplc} for 3D semantic segmentation.}
Figure~\ref{vocabulary_size} reports the mIoU and mAcc scores \rl{under different pre-training vocabulary sizes} $S \in \{10,12,15,150,170\}$, \zyja{on the ScanNet~\cite{scannet} and ScanNet200~\cite{scannet200} datasets}. Results show that the majority of the linguistic aspects of object attributes do not exhibit a notable enhancement as \rl{$S$ increases, reflected by} both mIoU and mAcc scores, aligning with our expectations. Some linguistic aspects of object attributes show \jy{relatively} low \rl{performances and exhibit} random jitters. Among \rl{the} eight linguistic aspects, the aspect \textit{material} illustrates an enhancement in mIoU and a marginal improvement in mAcc \rl{as $S$ increases}. This improvement can be attributed to the framework adopted by RegionPLC, which associates 3D objects with language through explicit visual image captioning models, providing detailed description\jy{s} of visual attributes like material and color for each 3D object. Therefore, as the vocabulary size $S$ increases, more objects are processed by the image captioning model to produce visual descriptions that ultimately improve the semantic segmentation results for the aspect \textit{material}.

This observation suggests that simply increasing \rl{the size of the} object vocabulary during training may not effectively enhance the generalization capability of OV-3D models. This limitation can be attributed to existing OV-3D benchmarks, like ScanNet~\cite{scannet}, ScanNet200~\cite{scannet200}, and ScanNet++~\cite{scannet++}, \rl{which} primarily focus on object classes and lack object-related attributes. 
\zyj{While increasing \rl{the size of} the object vocabulary during training can \ysqq{improves the OV-3D performance}, as demonstrated by \rl{the results from} PLA~\cite{ding2023pla} and RegionPLC~\cite{regionplc}, this approach is not suitable for the more challenging GOV-3D task. The significant performance gap between the two tasks cannot be resolved simply by transferring the OV-3D technique into GOV-3D.}

\subsection{The Impact of the Query Form}

\ysqq{In benchmark annotation, we generate queries linking attributes to object classes.}
An ideal query should contain an attribute name and the relation between the attribute and the corresponding object class. \ysqq{Table~\ref{query_form} shows the effect of using} a query template (\eg, ``\textit{this term is made of wood}'') versus \ysqq{a plain term} (\eg, ``\textit{wood}'') in GOV-3D. We evaluate the 3D instance segmentation of OpenMask3D~\cite{takmaz2023openmask3d}, SAI3D~\cite{sai3d}, MaskClustering~\cite{yan2024maskclustering}, and Open3DIS~\cite{nguyen2023open3dis}, \zyja{reporting the mean score of all attributes in OpenScan.} Note that, as expected, \ysqq{using the query template improves model performance, as shown by higher} AP, AP$_{50}$, and AP$_{25}$. SAI3D and MaskClustering \ysqq{are more} sensitive to query templates, while OpenMask3D and Open3DIS \ysqq{show greater robustness}. 

\ysqq{These results stem from the fact that VLMs, like} CLIP~\cite{clip}, \ysqq{struggle with attribute classification in GOV-3D when minor commonsense knowledge is required, as stated in~\cite{ye2023improving}}. \ysqq{Since most OV-3D models} rely on VLMs like CLIP~\cite{clip} for open-vocabulary comprehension, they inherit \ysqq{VLMs' commonsense limitations}. Thus, incorporating query templates 
\ysqq{that link attributes to object classes as commonsense knowledge improves OV-3D models' performances in GOV-3D.}

\vspace{-0.5mm}
\subsection{Qualitative Results}
\zyjn{We present qualitative results \rl{from} Open3DIS~\cite{nguyen2023open3dis} on our OpenScan benchmark. We evaluate Open3DIS across eight linguistic aspects, as shown in Figure~\ref{qualitative_supp}. It demonstrates that Open3DIS can comprehend specific linguistic aspects such as \textit{synonym} and \textit{material}. When exploring the \textit{affordance} aspect by querying ``\textit{keep food cold}'' \rl{for the target object, Open3DIS can successfully identify the ``\textit{refrigerator}'' as the target object}
but struggles to generate a correct 3D mask. Additionally, Open3DIS \rl{fails to} generate predictions for other linguistic aspects. These observations align with the quantitative results in Table~\ref{main_result}.}

\vspace{-0.5mm}

\subsection{Failure Cases Analysis}

\begin{figure}[tb] \centering
    \includegraphics[width=0.43\textwidth]{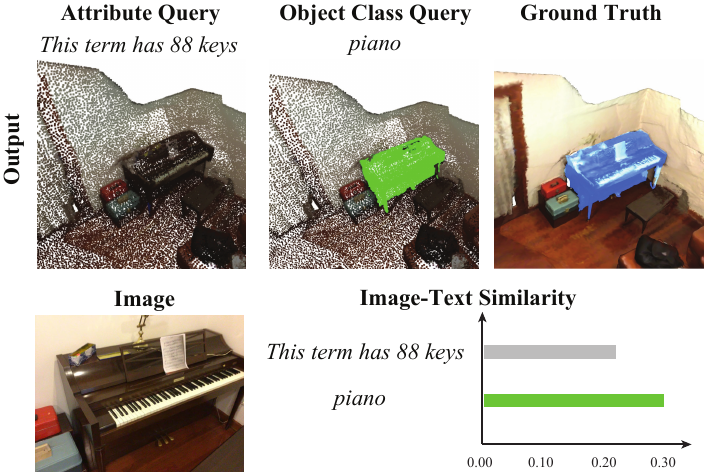}
    \vspace{-2mm}
    \caption{Visualization of the Open3DIS failure case with the corresponding CLIP image-text similarity scores.}
    \label{clip_sim}
    \vspace{-4mm}
\end{figure}

\zyjn{Figure~\ref{clip_sim} shows a failure case of Open3DIS~\cite{nguyen2023open3dis} when applied to GOV-3D, despite its strong performance on OV-3D. While Open3DIS correctly identifies the object class (\eg, ``\textit{piano}''), it fails to recognize the associated object attribute (\eg, ``\textit{this term has 88 keys}''). To investigate this discrepancy, we analyze the CLIP~\cite{clip} image-text similarity scores in Open3DIS, given that most OV-3D models rely on VLMs like CLIP~\cite{clip} for 3D predictions. Our analysis reveals that CLIP assigns lower image-text similarity scores to the object attribute compared to the object class, suggesting that its intrinsic attribute knowledge is limited. This observation demonstrates that GOV-3D presents greater challenges compared to OV-3D. A promising direction for GOV-3D involves integrating attribute knowledge into VLMs like CLIP, which may improve image-text alignment and enable more reliable predictions.
}
\section{Conclusion}
In this paper, we address the constraints of the classic Open-Vocabulary 3D Scene Understanding (OV-3D) task, which is limited in handling object attributes beyond object classes. We introduce a more challenging task, called Generalized Open-Vocabulary 3D Scene Understanding (GOV-3D), to comprehensively evaluate the generalization capability of OV-3D models. To facilitate research on the GOV-3D task, we construct a large-scale benchmark named OpenScan, \rl{which} consists of 341 attributes across 8 linguistic aspects. We systematically evaluate the OV-3D models on the OpenScan benchmark, revealing their challenges in understanding attributes beyond object classes. We also conduct experiments to investigate the impact of the \rl{pre-trained} vocabulary size and query form, demonstrating that the generalization ability can be enhanced by utilizing query templates rather than scaling up the vocabulary size during training. \zyjn{We further explore a promising direction for the GOV-3D task by integrating attribute knowledge into VLMs of the OV-3D models.} \rl{We believe} our OpenScan benchmark can facilitate future research on improving the generalization capability of OV-3D models.

{
    \small
    \bibliography{aaai2026}

\begin{thebibliography}{49}
\providecommand{\natexlab}[1]{#1}

\bibitem[{Armeni et~al.(2016)Armeni, Sener, Zamir, Jiang, Brilakis, Fischer, and Savarese}]{s3dis}
Armeni, I.; Sener, O.; Zamir, A.~R.; Jiang, H.; Brilakis, I.; Fischer, M.; and Savarese, S. 2016.
\newblock 3d semantic parsing of large-scale indoor spaces.
\newblock In \emph{Proceedings of the IEEE/CVF Conference on Computer Vision and Pattern Recognition}, 1534--1543.

\bibitem[{Baruch et~al.(2021)Baruch, Chen, Dehghan, Dimry, Feigin, Fu, Gebauer, Joffe, Kurz, Schwartz, and Shulman}]{dehghan2021arkitscenes}
Baruch, G.; Chen, Z.; Dehghan, A.; Dimry, T.; Feigin, Y.; Fu, P.; Gebauer, T.; Joffe, B.; Kurz, D.; Schwartz, A.; and Shulman, E. 2021.
\newblock {ARK}itScenes - A Diverse Real-World Dataset for 3D Indoor Scene Understanding Using Mobile {RGB}-D Data.
\newblock In \emph{Thirty-fifth Conference on Neural Information Processing Systems Datasets and Benchmarks Track (Round 1)}.

\bibitem[{Bianchi et~al.(2024)Bianchi, Carrara, Messina, Gennaro, and Falchi}]{fg-ovd}
Bianchi, L.; Carrara, F.; Messina, N.; Gennaro, C.; and Falchi, F. 2024.
\newblock The devil is in the fine-grained details: Evaluating open-vocabulary object detectors for fine-grained understanding.
\newblock In \emph{Proceedings of the IEEE/CVF Conference on Computer Vision and Pattern Recognition}, 22520--22529.

\bibitem[{Bojarski et~al.(2016)Bojarski, Del~Testa, Dworakowski, Firner, Flepp, Goyal, Jackel, Monfort, Muller, Zhang et~al.}]{autodrive}
Bojarski, M.; Del~Testa, D.; Dworakowski, D.; Firner, B.; Flepp, B.; Goyal, P.; Jackel, L.~D.; Monfort, M.; Muller, U.; Zhang, J.; et~al. 2016.
\newblock End to end learning for self-driving cars.
\newblock \emph{arXiv:1604.07316}.

\bibitem[{Chang et~al.(2017)Chang, Dai, Funkhouser, Halber, Niessner, Savva, Song, Zeng, and Zhang}]{Matterport3D}
Chang, A.; Dai, A.; Funkhouser, T.; Halber, M.; Niessner, M.; Savva, M.; Song, S.; Zeng, A.; and Zhang, Y. 2017.
\newblock {Matterport3D}: Learning from {RGB-D} Data in Indoor Environments.
\newblock \emph{International Conference on 3D Vision}.

\bibitem[{Chen, Chang, and Nie{\ss}ner(2020)}]{chen2020scanrefer}
Chen, D.~Z.; Chang, A.~X.; and Nie{\ss}ner, M. 2020.
\newblock Scanrefer: 3d object localization in rgb-d scans using natural language.
\newblock In \emph{European Conference on Computer Vision}, 202--221. Springer.

\bibitem[{Chiang et~al.(2023)Chiang, Li, Lin, Sheng, Wu, Zhang, Zheng, Zhuang, Zhuang, Gonzalez, Stoica, and Xing}]{vicuna2023}
Chiang, W.-L.; Li, Z.; Lin, Z.; Sheng, Y.; Wu, Z.; Zhang, H.; Zheng, L.; Zhuang, S.; Zhuang, Y.; Gonzalez, J.~E.; Stoica, I.; and Xing, E.~P. 2023.
\newblock Vicuna: An Open-Source Chatbot Impressing GPT-4 with 90\%* ChatGPT Quality.

\bibitem[{Choy, Gwak, and Savarese(2019)}]{MinkowskiNet}
Choy, C.; Gwak, J.; and Savarese, S. 2019.
\newblock 4d spatio-temporal convnets: Minkowski convolutional neural networks.
\newblock In \emph{Proceedings of the IEEE/CVF Conference on Computer Vision and Pattern Recognition}, 3075--3084.

\bibitem[{Cordts et~al.(2016)Cordts, Omran, Ramos, Rehfeld, Enzweiler, Benenson, Franke, Roth, and Schiele}]{cordts2016cityscapes}
Cordts, M.; Omran, M.; Ramos, S.; Rehfeld, T.; Enzweiler, M.; Benenson, R.; Franke, U.; Roth, S.; and Schiele, B. 2016.
\newblock The cityscapes dataset for semantic urban scene understanding.
\newblock In \emph{Proceedings of the IEEE Conference on Computer Vision and Pattern Recognition}, 3213--3223.

\bibitem[{Dai et~al.(2017)Dai, Chang, Savva, Halber, Funkhouser, and Nie{\ss}ner}]{scannet}
Dai, A.; Chang, A.~X.; Savva, M.; Halber, M.; Funkhouser, T.; and Nie{\ss}ner, M. 2017.
\newblock Scannet: Richly-annotated 3d reconstructions of indoor scenes.
\newblock In \emph{Proceedings of the IEEE Conference on Computer Vision and Pattern Recognition}, 5828--5839.

\bibitem[{Delitzas et~al.(2024)Delitzas, Takmaz, Tombari, Sumner, Pollefeys, and Engelmann}]{delitzas2024scenefun3d}
Delitzas, A.; Takmaz, A.; Tombari, F.; Sumner, R.; Pollefeys, M.; and Engelmann, F. 2024.
\newblock Scenefun3d: Fine-grained functionality and affordance understanding in 3d scenes.
\newblock In \emph{Proceedings of the IEEE/CVF Conference on Computer Vision and Pattern Recognition}, 14531--14542.

\bibitem[{Ding et~al.(2023)Ding, Yang, Xue, Zhang, Bai, and Qi}]{ding2023pla}
Ding, R.; Yang, J.; Xue, C.; Zhang, W.; Bai, S.; and Qi, X. 2023.
\newblock PLA: Language-Driven Open-Vocabulary 3D Scene Understanding.
\newblock In \emph{Proceedings of the IEEE/CVF Conference on Computer Vision and Pattern Recognition}, 7010--7019.

\bibitem[{Ester et~al.(1996)Ester, Kriegel, Sander, and Xu}]{dbscan}
Ester, M.; Kriegel, H.-P.; Sander, J.; and Xu, X. 1996.
\newblock A density-based algorithm for discovering clusters in large spatial databases with noise.
\newblock In \emph{Proceedings of the Second International Conference on Knowledge Discovery and Data Mining}, KDD'96, 226–231. AAAI Press.

\bibitem[{Everingham et~al.(2015)Everingham, Eslami, Van~Gool, Williams, Winn, and Zisserman}]{everingham2015pascal}
Everingham, M.; Eslami, S.~A.; Van~Gool, L.; Williams, C.~K.; Winn, J.; and Zisserman, A. 2015.
\newblock The pascal visual object classes challenge: A retrospective.
\newblock \emph{International Journal of Computer Vision}, 111: 98--136.

\bibitem[{Ghiasi et~al.(2022)Ghiasi, Gu, Cui, and Lin}]{openseg}
Ghiasi, G.; Gu, X.; Cui, Y.; and Lin, T.-Y. 2022.
\newblock Scaling open-vocabulary image segmentation with image-level labels.
\newblock In \emph{European Conference on Computer Vision}, 540--557. Springer.

\bibitem[{Graham, Engelcke, and Van Der~Maaten(2018)}]{sparse-convolution}
Graham, B.; Engelcke, M.; and Van Der~Maaten, L. 2018.
\newblock 3d semantic segmentation with submanifold sparse convolutional networks.
\newblock In \emph{Proceedings of the IEEE/CVF Conference on Computer Vision and Pattern Recognition}, 9224--9232.

\bibitem[{Gu et~al.(2022)Gu, Lin, Kuo, and Cui}]{gu2021open}
Gu, X.; Lin, T.-Y.; Kuo, W.; and Cui, Y. 2022.
\newblock Open-vocabulary Object Detection via Vision and Language Knowledge Distillation.
\newblock In \emph{International Conference on Learning Representations}.

\bibitem[{Gupta, Dollar, and Girshick(2019)}]{lvis}
Gupta, A.; Dollar, P.; and Girshick, R. 2019.
\newblock Lvis: A dataset for large vocabulary instance segmentation.
\newblock In \emph{Proceedings of the IEEE/CVF Conference on Computer Vision and Pattern Recognition}, 5356--5364.

\bibitem[{Huang et~al.(2025)Huang, Li, Qi, Yan, and Yang}]{huang2025reason3d}
Huang, K.-C.; Li, X.; Qi, L.; Yan, S.; and Yang, M.-H. 2025.
\newblock Reason3d: Searching and reasoning 3d segmentation via large language model.
\newblock In \emph{International Conference on 3D Vision 2025}.

\bibitem[{Huang et~al.(2024)Huang, Wu, Chen, Zhao, Zhu, and Lasenby}]{huang2024openins3d}
Huang, Z.; Wu, X.; Chen, X.; Zhao, H.; Zhu, L.; and Lasenby, J. 2024.
\newblock Openins3d: Snap and lookup for 3d open-vocabulary instance segmentation.
\newblock In \emph{European Conference on Computer Vision}, 169--185. Springer.

\bibitem[{Kirillov et~al.(2023)Kirillov, Mintun, Ravi, Mao, Rolland, Gustafson, Xiao, Whitehead, Berg, Lo, Dollar, and Girshick}]{sam}
Kirillov, A.; Mintun, E.; Ravi, N.; Mao, H.; Rolland, C.; Gustafson, L.; Xiao, T.; Whitehead, S.; Berg, A.~C.; Lo, W.-Y.; Dollar, P.; and Girshick, R. 2023.
\newblock Segment Anything.
\newblock In \emph{Proceedings of the IEEE/CVF International Conference on Computer Vision}, 4015--4026.

\bibitem[{Li et~al.(2024)Li, Zhang, Sun, Zou, Liu, Li, Yang, Zhang, and Gao}]{semantic-sam}
Li, F.; Zhang, H.; Sun, P.; Zou, X.; Liu, S.; Li, C.; Yang, J.; Zhang, L.; and Gao, J. 2024.
\newblock Segment and recognize anything at any granularity.
\newblock In \emph{European Conference on Computer Vision}, 467--484. Springer.

\bibitem[{Lin et~al.(2014)Lin, Maire, Belongie, Hays, Perona, Ramanan, Doll{\'a}r, and Zitnick}]{coco}
Lin, T.-Y.; Maire, M.; Belongie, S.; Hays, J.; Perona, P.; Ramanan, D.; Doll{\'a}r, P.; and Zitnick, C.~L. 2014.
\newblock Microsoft coco: Common objects in context.
\newblock In \emph{European Conference on Computer Vision}, 740--755. Springer.

\bibitem[{Liu et~al.(2024)Liu, Zeng, Ren, Li, Zhang, Yang, Jiang, Li, Yang, Su et~al.}]{liu2024groundingdino}
Liu, S.; Zeng, Z.; Ren, T.; Li, F.; Zhang, H.; Yang, J.; Jiang, Q.; Li, C.; Yang, J.; Su, H.; et~al. 2024.
\newblock Grounding dino: Marrying dino with grounded pre-training for open-set object detection.
\newblock In \emph{European Conference on Computer Vision}, 38--55. Springer.

\bibitem[{Lu et~al.(2023)Lu, Chang, Jing, Boularias, and Bekris}]{lu2023ovir3d}
Lu, S.; Chang, H.; Jing, E.~P.; Boularias, A.; and Bekris, K. 2023.
\newblock Ovir-3d: Open-vocabulary 3d instance retrieval without training on 3d data.
\newblock In \emph{Conference on Robot Learning}, 1610--1620. PMLR.

\bibitem[{Lyu et~al.(2024)Lyu, Lin, Wang, Mao, Chen, Xu, Huang, Zhu, Lin, and Pang}]{lyu2024mmscan}
Lyu, R.; Lin, J.; Wang, T.; Mao, X.; Chen, Y.; Xu, R.; Huang, H.; Zhu, C.; Lin, D.; and Pang, J. 2024.
\newblock Mmscan: A multi-modal 3d scene dataset with hierarchical grounded language annotations.
\newblock \emph{Advances in Neural Information Processing Systems}, 37: 50898--50924.

\bibitem[{Ngo, Hua, and Nguyen(2023)}]{isbnet}
Ngo, T.~D.; Hua, B.-S.; and Nguyen, K. 2023.
\newblock ISBNet: A 3D Point Cloud Instance Segmentation Network With Instance-Aware Sampling and Box-Aware Dynamic Convolution.
\newblock In \emph{Proceedings of the IEEE/CVF Conference on Computer Vision and Pattern Recognition}, 13550--13559.

\bibitem[{Nguyen et~al.(2024)Nguyen, Ngo, Kalogerakis, Gan, Tran, Pham, and Nguyen}]{nguyen2023open3dis}
Nguyen, P.; Ngo, T.~D.; Kalogerakis, E.; Gan, C.; Tran, A.; Pham, C.; and Nguyen, K. 2024.
\newblock Open3dis: Open-vocabulary 3d instance segmentation with 2d mask guidance.
\newblock In \emph{Proceedings of the IEEE/CVF Conference on Computer Vision and Pattern Recognition}, 4018--4028.

\bibitem[{Peng et~al.(2023)Peng, Genova, Jiang, Tagliasacchi, Pollefeys, Funkhouser et~al.}]{peng2023openscene}
Peng, S.; Genova, K.; Jiang, C.; Tagliasacchi, A.; Pollefeys, M.; Funkhouser, T.; et~al. 2023.
\newblock Openscene: 3d scene understanding with open vocabularies.
\newblock In \emph{Proceedings of the IEEE/CVF Conference on Computer Vision and Pattern Recognition}, 815--824.

\bibitem[{Qi et~al.(2023)Qi, Kuen, Shen, Gu, Li, Guo, Jia, Lin, and Yang}]{cropformer}
Qi, L.; Kuen, J.; Shen, T.; Gu, J.; Li, W.; Guo, W.; Jia, J.; Lin, Z.; and Yang, M.-H. 2023.
\newblock High Quality Entity Segmentation.
\newblock In \emph{2023 IEEE/CVF International Conference on Computer Vision}, 4024--4033.

\bibitem[{Radford et~al.(2021)Radford, Kim, Hallacy, Ramesh, Goh, Agarwal, Sastry, Askell, Mishkin, Clark et~al.}]{clip}
Radford, A.; Kim, J.~W.; Hallacy, C.; Ramesh, A.; Goh, G.; Agarwal, S.; Sastry, G.; Askell, A.; Mishkin, P.; Clark, J.; et~al. 2021.
\newblock Learning transferable visual models from natural language supervision.
\newblock In \emph{International Conference on Machine Learning}, 8748--8763. PMLR.

\bibitem[{Ramanathan et~al.(2023)Ramanathan, Kalia, Petrovic, Wen, Zheng, Guo, Wang, Marquez, Kovvuri, Kadian et~al.}]{ramanathan2023paco}
Ramanathan, V.; Kalia, A.; Petrovic, V.; Wen, Y.; Zheng, B.; Guo, B.; Wang, R.; Marquez, A.; Kovvuri, R.; Kadian, A.; et~al. 2023.
\newblock Paco: Parts and attributes of common objects.
\newblock In \emph{Proceedings of the IEEE/CVF Conference on Computer Vision and Pattern Recognition}, 7141--7151.

\bibitem[{Ren et~al.(2024)Ren, Liu, Zeng, Lin, Li, Cao, Chen, Huang, Chen, Yan et~al.}]{groundedsam}
Ren, T.; Liu, S.; Zeng, A.; Lin, J.; Li, K.; Cao, H.; Chen, J.; Huang, X.; Chen, Y.; Yan, F.; et~al. 2024.
\newblock Grounded sam: Assembling open-world models for diverse visual tasks.
\newblock \emph{arXiv:2401.14159}.

\bibitem[{Rozenberszki et~al.(2022)Rozenberszki, Litany, Dai, and Dai}]{scannet200}
Rozenberszki, D.; Litany, O.; Dai, A.; and Dai, A. 2022.
\newblock Language-Grounded Indoor 3D Semantic Segmentation in the Wild.
\newblock In \emph{Proceedings of the European Conference on Computer Vision}.

\bibitem[{Schult et~al.(2023)Schult, Engelmann, Hermans, Litany, Tang, and Leibe}]{mask3d}
Schult, J.; Engelmann, F.; Hermans, A.; Litany, O.; Tang, S.; and Leibe, B. 2023.
\newblock Mask3D: Mask Transformer for 3D Semantic Instance Segmentation.
\newblock In \emph{2023 IEEE International Conference on Robotics and Automation}, 8216--8223.

\bibitem[{Speer, Chin, and Havasi(2017)}]{speer2017conceptnet}
Speer, R.; Chin, J.; and Havasi, C. 2017.
\newblock Conceptnet 5.5: An open multilingual graph of general knowledge.
\newblock In \emph{Proceedings of the AAAI Conference on Artificial Intelligence}, volume~31.

\bibitem[{Straub et~al.(2019)Straub, Whelan, Ma, Chen, Wijmans, Green, Engel, Mur-Artal, Ren, Verma et~al.}]{replica}
Straub, J.; Whelan, T.; Ma, L.; Chen, Y.; Wijmans, E.; Green, S.; Engel, J.~J.; Mur-Artal, R.; Ren, C.; Verma, S.; et~al. 2019.
\newblock The Replica dataset: A digital replica of indoor spaces.
\newblock \emph{arXiv:1906.05797}.

\bibitem[{Takmaz et~al.(2023)Takmaz, Fedele, Sumner, Pollefeys, Tombari, and Engelmann}]{takmaz2023openmask3d}
Takmaz, A.; Fedele, E.; Sumner, R.~W.; Pollefeys, M.; Tombari, F.; and Engelmann, F. 2023.
\newblock {OpenMask3D: Open-Vocabulary 3D Instance Segmentation}.
\newblock In \emph{Advances in Neural Information Processing Systems}.

\bibitem[{Xu et~al.(2024)Xu, Shi, Tu, Zhou, Liang, and Bai}]{UniSeg3D}
Xu, W.; Shi, C.; Tu, S.; Zhou, X.; Liang, D.; and Bai, X. 2024.
\newblock A Unified Framework for 3D Scene Understanding.
\newblock In \emph{Advances in Neural Information Processing Systems}.

\bibitem[{Yan et~al.(2024)Yan, Zhang, Zhu, and Wang}]{yan2024maskclustering}
Yan, M.; Zhang, J.; Zhu, Y.; and Wang, H. 2024.
\newblock Maskclustering: View consensus based mask graph clustering for open-vocabulary 3d instance segmentation.
\newblock In \emph{Proceedings of the IEEE/CVF Conference on Computer Vision and Pattern Recognition}, 28274--28284.

\bibitem[{Yang et~al.(2024)Yang, Ding, Deng, Wang, and Qi}]{regionplc}
Yang, J.; Ding, R.; Deng, W.; Wang, Z.; and Qi, X. 2024.
\newblock Regionplc: Regional point-language contrastive learning for open-world 3d scene understanding.
\newblock In \emph{Proceedings of the IEEE/CVF Conference on Computer Vision and Pattern Recognition}, 19823--19832.

\bibitem[{Yao et~al.(2024)Yao, Liu, Zhao, Zhang, Liao, Fang, Lee, and Wang}]{ovdeval}
Yao, Y.; Liu, P.; Zhao, T.; Zhang, Q.; Liao, J.; Fang, C.; Lee, K.; and Wang, Q. 2024.
\newblock How to Evaluate the Generalization of Detection? A Benchmark for Comprehensive Open-Vocabulary Detection.
\newblock In \emph{Proceedings of the AAAI Conference on Artificial Intelligence}, volume~38, 6630--6638.

\bibitem[{Ye et~al.(2023)Ye, Xie, Chen, Xu, Yuan, Zhu, and Liao}]{ye2023improving}
Ye, S.; Xie, Y.; Chen, D.; Xu, Y.; Yuan, L.; Zhu, C.; and Liao, J. 2023.
\newblock Improving commonsense in vision-language models via knowledge graph riddles.
\newblock In \emph{Proceedings of the IEEE/CVF Conference on Computer Vision and Pattern Recognition}, 2634--2645.

\bibitem[{Yeshwanth et~al.(2023)Yeshwanth, Liu, Nie{\ss}ner, and Dai}]{scannet++}
Yeshwanth, C.; Liu, Y.-C.; Nie{\ss}ner, M.; and Dai, A. 2023.
\newblock Scannet++: A high-fidelity dataset of 3d indoor scenes.
\newblock In \emph{Proceedings of the IEEE/CVF International Conference on Computer Vision}, 12--22.

\bibitem[{Yin et~al.(2024)Yin, Liu, Xiao, Cohen-Or, Huang, and Chen}]{sai3d}
Yin, Y.; Liu, Y.; Xiao, Y.; Cohen-Or, D.; Huang, J.; and Chen, B. 2024.
\newblock Sai3d: Segment any instance in 3d scenes.
\newblock In \emph{Proceedings of the IEEE/CVF Conference on Computer Vision and Pattern Recognition}, 3292--3302.

\bibitem[{Zeng et~al.(2018)Zeng, Song, Welker, Lee, Rodriguez, and Funkhouser}]{robotic}
Zeng, A.; Song, S.; Welker, S.; Lee, J.; Rodriguez, A.; and Funkhouser, T. 2018.
\newblock Learning synergies between pushing and grasping with self-supervised deep reinforcement learning.
\newblock In \emph{2018 IEEE/RSJ International Conference on Intelligent Robots and Systems}, 4238--4245. IEEE.

\bibitem[{Zhao, Lin, and Lau(2025)}]{hcma}
Zhao, Y.; Lin, J.; and Lau, R.~W. 2025.
\newblock Hierarchical Cross-Modal Alignment for Open-Vocabulary 3D Object Detection.
\newblock In \emph{Proceedings of the AAAI Conference on Artificial Intelligence}, volume~39, 10501--10509.

\bibitem[{Zhong et~al.(2022)Zhong, Yang, Zhang, Li, Codella, Li, Zhou, Dai, Yuan, Li et~al.}]{regionclip}
Zhong, Y.; Yang, J.; Zhang, P.; Li, C.; Codella, N.; Li, L.~H.; Zhou, L.; Dai, X.; Yuan, L.; Li, Y.; et~al. 2022.
\newblock Regionclip: Region-based language-image pretraining.
\newblock In \emph{Proceedings of the IEEE/CVF Conference on Computer Vision and Pattern Recognition}, 16793--16803.

\bibitem[{Zhou et~al.(2019)Zhou, Zhao, Puig, Xiao, Fidler, Barriuso, and Torralba}]{ade20k}
Zhou, B.; Zhao, H.; Puig, X.; Xiao, T.; Fidler, S.; Barriuso, A.; and Torralba, A. 2019.
\newblock Semantic understanding of scenes through the ade20k dataset.
\newblock \emph{International Journal of Computer Vision}, 127: 302--321.

\end{thebibliography}
}

\clearpage

\clearpage
\setcounter{page}{1}
\setcounter{secnumdepth}{2}

\setcounter{section}{0}
\setcounter{figure}{0}
\setcounter{table}{0}
\renewcommand{\thesection}{\Alph{section}}
\renewcommand{\thefigure}{\Alph{figure}}
\renewcommand{\thetable}{\Alph{table}}

\centerline{\LARGE{\textbf{Supplementary Material}}}

\vspace{0.3cm}
In this supplementary material, we provide more experimental results and benchmark details:

\begin{itemize}
    \item Sec.~\ref{web_interface}: Web interface for manual annotation.
    \item Sec.~\ref{more_exp_detail}: Implementation details.
    \item Sec.~\ref{add_exp_results}: Additional experimental results.
    \item Sec.~\ref{add_benchmark_detail}: Additional benchmark details.
    \item Sec.~\ref{add_related}: Additional related work.    
    \item Sec.~\ref{limit_future}: Limitations and future work.
    \item Sec.~\ref{broader_impact}: Broader impact.
\end{itemize}

\section{Web Interface for Manual Annotation}
\label{web_interface}
\begin{figure}[t]
    \centering
    \includegraphics[width=0.46\textwidth]{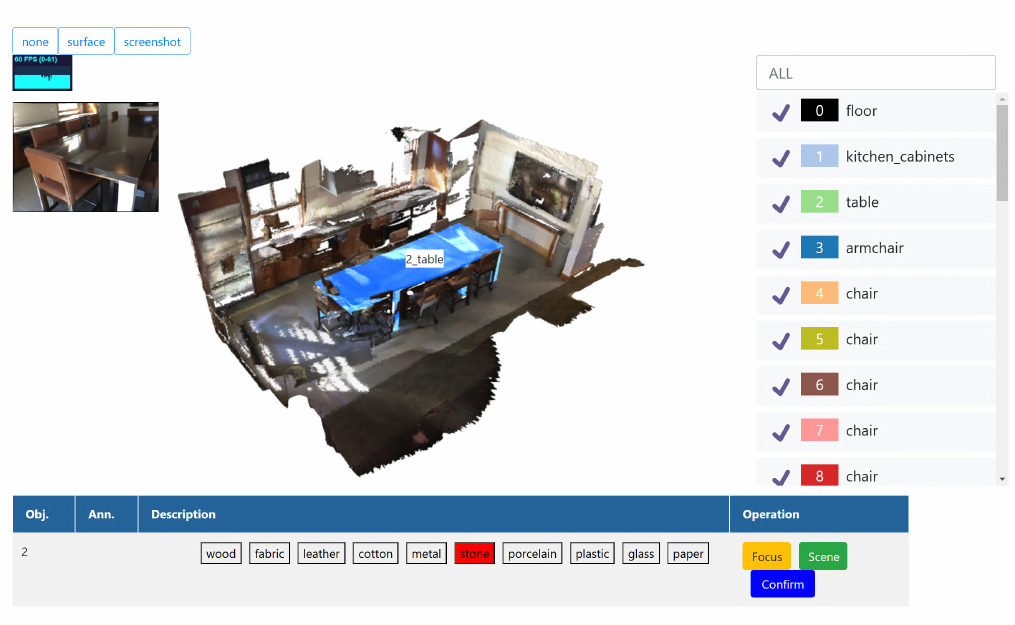}
    \caption{Web interface for manual annotation that allows users to view the 3D scene from multiple viewpoints and select the target object by clicking.}
    \label{interface}
\end{figure}

We implement a web interface for manual annotation of the visual linguistic aspect (\eg, \textit{material}), as shown in Figure~\ref{interface}. Annotators are shown an interactive 3D mesh of a scene, a list of target objects, and a list of attributes. Users can control the 3D mesh from different viewpoints interactively by rotating, zooming in, zooming out, and panning to observe the scene from various viewpoints. When users select a 3D mesh by clicking the mouse in the scene, the target object will be highlighted and the corresponding object ID and object class will be displayed. The annotation process requires annotators to first select a target object in the 3D mesh (\eg, table of ID 2) and then select a primary attribute that belongs to the target object (\eg, stone). Finally, annotators click the confirm button to submit and store the annotations. Once the selected objects are annotated and confirmed, the corresponding object in the object list will show a check mark symbol. Additionally, to address visual ambiguity issues in 3D object appearance, our annotation process allows users to review the scene’s video sequences, providing contextual visual cues to resolve uncertainties about target objects’ visual attributes during annotation.

\section{Implementation Details}
\label{more_exp_detail}
\subsection{Open-Vocabulary 3D Scene Understanding (OV-3D) Baselines}
We report implementation details of the OV-3D models~\cite{takmaz2023openmask3d, sai3d, yan2024maskclustering, nguyen2023open3dis, peng2023openscene, ding2023pla, regionplc} as follows:
\noindent\textbf{OpenMask3D.}
In the class-agnostic mask proposal module, we employ the Mask3D~\cite{mask3d} architecture trained on the ScanNet200~\cite{scannet200} training set. For 2D mask proposals, we use SAM~\cite{sam} with ViT-H as the backbone. We utilize the pre-trained CLIP~\cite{clip} visual encoder of ViT-L/14 at a 336 pixel resolution to extract image features with 768 dimensions. We set the number of queries to 150, following the implementation of OpenMask3D~\cite{takmaz2023openmask3d} and Mask3D~\cite{mask3d}, to ensure a sufficient number of mask proposals for the GOV-3D task.

\noindent\textbf{SAI3D.} 
We employ Semantic-SAM~\cite{semantic-sam} with Swin-L as the backbone to generate 2D mask proposals. The number of queries is set to 150 to ensure sufficient mask proposals for the GOV-3D task.

\noindent\textbf{MaskClustering.} 
We utilize CropFormer~\cite{cropformer} as a 2D mask predictor. For 2D mask proposals, we use CLIP~\cite{clip} visual encoder of ViT-H/14 to extract image features. We follow MaskClustering~\cite{yan2024maskclustering} to adopt the post-processing approach from OVIR-3D~\cite{lu2023ovir3d} to refine the output 3D instances. Specifically, we employ the DBSCAN~\cite{dbscan} algorithm to partition disconnected point clusters.

\noindent\textbf{Open3DIS.} 
We utilize the class-agnostic 3D proposal network ISBNet~\cite{isbnet} trained on the ScanNet200~\cite{scannet200} training set as 3D proposal. We employ the 2D-Guided-3D Instance Proposal Module in Open3DIS~\cite{nguyen2023open3dis}. For 2D mask proposals, we adopt Grounded-SAM~\cite{groundedsam} as 2D segmentor, which incorporates a Swin-T-based Grounding-DINO~\cite{liu2024groundingdino} decoder and SAM~\cite{sam} with ViT-H as the backbone.

\noindent\textbf{OpenScene.}
We employ OpenSeg~\cite{openseg} for image feature extraction and a 2D-3D ensemble model in OpenScene~\cite{peng2023openscene}. We utilize MinkowskiNet18A~\cite{MinkowskiNet} as the 3D backbone during 3D distillation.

\noindent\textbf{PLA.}
We utilize a model trained on the ScanNet~\cite{scannet} partition of B15/N4, where B15/N4 indicates 15 base and 4 novel categories. We adopt a SparseUNet16 architecture based on sparse convolutions UNet~\cite{sparse-convolution} as our 3D encoder for semantic segmentation and integrate the CLIP~\cite{clip} text encoder as the final classifier.

\noindent\textbf{RegionPLC.}
We utilize a model trained on the ScanNet~\cite{scannet} partition of B15/N4, where B15/N4 represents 15 base and 4 novel categories. We employ a sparse-convolution-based UNet~\cite{sparse-convolution} of SparseUNet16 as the 3D encoder for semantic segmentation, leveraging the CLIP~\cite{clip} text encoder as the final classifier.

\begin{figure*}[ht] \centering
    \includegraphics[width=0.98\textwidth]{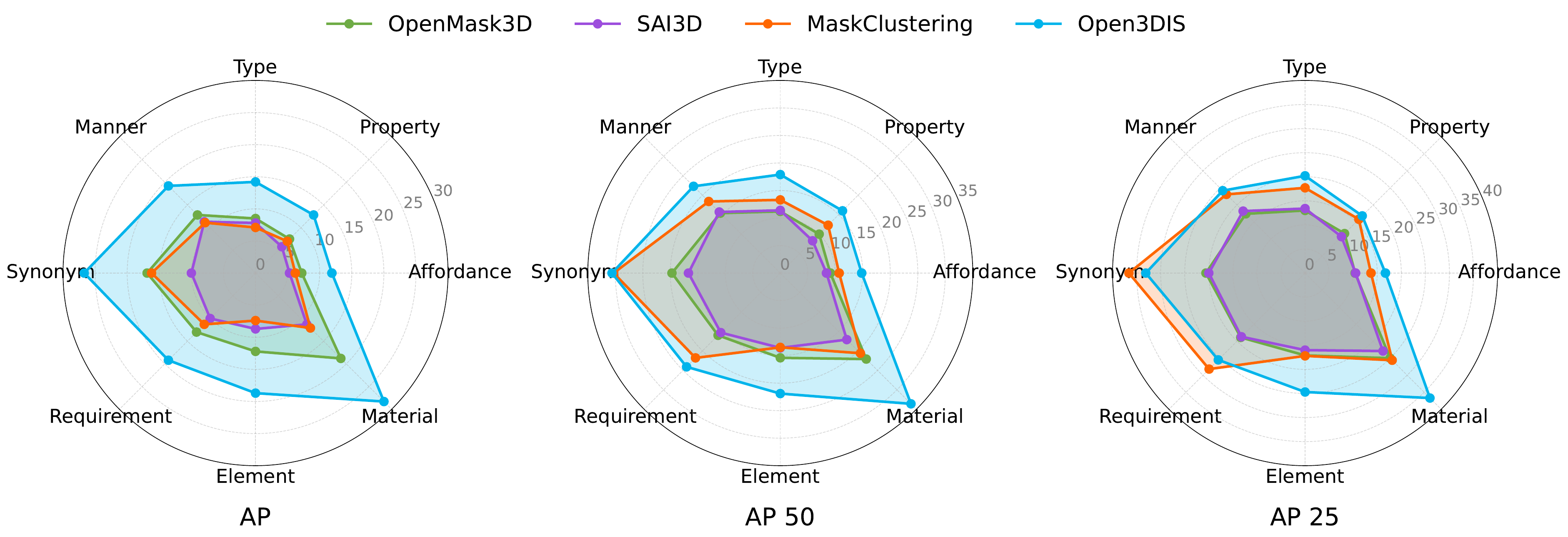}
    \vspace{2mm}
    \caption{Radar charts of AP, AP$_{50}$, and AP$_{25}$ result\jy{s} for \jy{eight} linguistic aspects on our OpenScan benchmark.}
    \label{radia}
    \vspace{2mm}
\end{figure*}

\begin{table*}[h]
\renewcommand\arraystretch{1.1}
    \resizebox{0.98\textwidth}{!}
    {
    \begin{tabular}{l|cccccccc|c}
        \toprule
     \textbf{Method} & \textbf{Affordance} &  \textbf{Property} & \textbf{Type} & \textbf{Manner} & \textbf{Synonym} & \textbf{Requirement} & \textbf{Element} & \textbf{Material} & \textbf{Mean} \\
        \midrule
        \multicolumn{10}{c}{\textbf{AP}} \\
        \midrule
        OpenMask3D~\cite{takmaz2023openmask3d} & 7.7 & 8.2 & 8.7 & 14.2 & 16.3 & 12.0 & 10.0 & 14.7 & 9.7 \\
        SAI3D~\cite{sai3d} & 4.0 & 5.5 & 7.4 & 10.6 & 8.6 & 8.8 & 7.3 & 8.3 & 6.7 \\
        MaskClustering~\cite{yan2024maskclustering} & 6.0 & 6.2 & 4.1 & 8.0 & 11.7 & 10.3 & 9.3 & 8.1 & 6.8 \\
        Open3DIS~\cite{nguyen2023open3dis} & \textbf{11.5} & \textbf{17.6} & \textbf{14.5} & \textbf{18.5} & \textbf{27.7} & \textbf{18.6} & \textbf{16.6} & \textbf{23.4} & \textbf{15.6} \\
        \midrule
        \multicolumn{10}{c}{\textbf{AP$_{50}$}} \\
        \midrule
        OpenMask3D~\cite{takmaz2023openmask3d} & 9.6 & 10.7 & 11.2 & 18.1 & 18.4 & 15.2 & 12.8 & 17.8 & 12.2 \\
        SAI3D~\cite{sai3d} & 6.4 & 8.2 & 10.6 & 14.9 & 13.0 & 13.8 & 11.5 & 13.3 & 10.1 \\
        MaskClustering~\cite{yan2024maskclustering} & 10.3 & 11.5 & 7.2 & 13.5 & 23.1 & 18.7 & 16.1 & 14.0 & 12.0 \\
        Open3DIS~\cite{nguyen2023open3dis} & \textbf{14.3} & \textbf{22.2} & \textbf{18.3} & \textbf{22.6} & \textbf{32.9} & \textbf{23.1} & \textbf{19.7} & \textbf{28.5} & \textbf{19.2} \\
        \midrule
        \multicolumn{10}{c}{\textbf{AP$_{25}$}} \\
        \midrule
        OpenMask3D~\cite{takmaz2023openmask3d} & 10.8 & 12.4 & 13.9 & 20.4 & 18.9 & 18.0 & 14.4 & 20.5 & 14.1 \\
        SAI3D~\cite{sai3d}  & 8.5 & 11.1 & 13.1 & 18.6 & 15.9 & 17.5 & 14.7 & 18.6 & 12.8 \\
        MaskClustering~\cite{yan2024maskclustering} & 12.3 & 13.5 & 9.4 & 16.9 & 25.1 & 24.1 & 19.0 & 18.1 & 14.6 \\
        Open3DIS~\cite{nguyen2023open3dis} & \textbf{16.2} & \textbf{24.0} & \textbf{20.2} & \textbf{24.8} & \textbf{35.9} & \textbf{24.9} & \textbf{22.7} & \textbf{31.9} & \textbf{21.3} \\
        \toprule
    \end{tabular}
    }
    \vspace{1mm}
    \caption{3D instance segmentation results without query template on our OpenScan benchmark.} 
\label{main_result_w/o_template}
\end{table*}

\subsection{Evaluation Protocol}
We evaluate the OV-3D baselines on 312 scenes following the validation split of ScanNet200~\cite{scannet200} dataset. Our evaluation includes all 341 attributes across 8 linguistic aspects.

\section{Additional Experimental Results}
\label{add_exp_results}
\vspace{1mm}

\subsection{Results of Radar Charts.}

Figure~\ref{radia} presents radar charts of the main results on our OpenScan benchmark. The charts compare performance across AP, AP$_{50}$, and AP$_{25}$ for OpenMask3D~\cite{takmaz2023openmask3d}, SAI3D~\cite{sai3d}, MaskClustering~\cite{yan2024maskclustering}, and Open3DIS~\cite{nguyen2023open3dis}. Our results demonstrate that Open3DIS achieves the strongest performance across eight linguistic aspects, particularly in terms of AP and AP$_{50}$. Meanwhile, MaskClustering exhibits competitive performance in AP$_{50}$ and AP$_{25}$, with notable strengths in the \textit{synonym} and \textit{requirement} aspects.

\begin{table*}[h]
    \centering
    \resizebox{0.98\textwidth}{!}
    {
    \begin{tabular}{l|cccccccccc|c}
        \toprule
      \textbf{Method}  & \textbf{Wood} &  \textbf{Fabric} & \textbf{Leather} &  \textbf{Cotton} & \textbf{Metal} & \textbf{Stone} & \textbf{Porcelain} & \textbf{Plastic} & \textbf{Glass} & \textbf{Paper} & \textbf{Mean} \\
        \midrule
        \multicolumn{12}{c}{\textbf{AP}}
        \\
        \midrule
        OpenMask3D~\cite{takmaz2023openmask3d} & 19.1 & 12.7 & 28.0 & 26.5 & 9.1 & 0.1 & 41.8 & 16.9 & 23.1 & 10.7 & 18.8 \\
        SAI3D~\cite{sai3d} & 13.1 & 10.3 & 19.2 & 5.6 & 6.5 & 0.1 & 19.7 & 12.8 & 14.9 & 10.4 & 11.3 \\
        MaskClustering~\cite{yan2024maskclustering} & 12.8 & 18.7 & 26.3 & 10.4 & 6.3 & 0.3 & 25.1 & 13.2 & 4.8 & 3.5 & 12.1 \\
        Open3DIS~\cite{nguyen2023open3dis} & \textbf{32.9} & \textbf{27.1} & \textbf{35.4} & \textbf{33.0} & \textbf{24.6} & \textbf{2.4} & \textbf{43.1} & \textbf{33.8} & \textbf{29.9} & \textbf{20.9} & \textbf{28.3} \\
        \midrule
        \multicolumn{12}{c}{\textbf{AP}$_{50}$}
        \\
        \midrule
        OpenMask3D~\cite{takmaz2023openmask3d} & 23.9 & 16.0 & 30.4 & 30.3 & 11.6 & 0.1 & 44.5 & 20.0 & 29.7 & 15.2 & 22.1\\
        SAI3D~\cite{sai3d} & 19.6 & 15.7 & 26.9 & 9.5 & 10.2 & 0.1 & 31.8 & 18.4 & 22.6 & 16.2 & 17.1 \\
        MaskClustering~\cite{yan2024maskclustering} & 23.8 & 31.6 & \textbf{41.0} & 19.7 & 12.0 & 0.5 & 37.0 & 23.2 & 9.3 & 7.6 & 20.6 \\
        Open3DIS~\cite{nguyen2023open3dis} & \textbf{40.9} & \textbf{32.7} & 39.0 & \textbf{38.6} & \textbf{30.7} & \textbf{3.5} & \textbf{46.3} & \textbf{38.8} & \textbf{37.6} & \textbf{27.8} & \textbf{33.6}\\
        \midrule
        \multicolumn{12}{c}{\textbf{AP}$_{25}$}
        \\
        \midrule  
        OpenMask3D~\cite{takmaz2023openmask3d} & 27.4 & 19.1 & 32.5 & 33.5 & 13.3 & 0.1 & 47.0 & 21.7 & 34.8 & 21.4 & 25.0\\
        SAI3D~\cite{sai3d} & 26.4 & 20.5 & 32.2 & 18.5 & 13.7 & 0.2 & 41.0 & 23.4 & 31.4 & 22.0 & 22.9 \\
        MaskClustering~\cite{yan2024maskclustering} & 31.3 & \textbf{38.4} & \textbf{49.7} & 23.0 & 16.7 & 0.6 & 41.6 & 28.2 & 16.1 & 10.9 & 25.6\\
        Open3DIS~\cite{nguyen2023open3dis} & \textbf{44.7} & 35.3 & 42.6 & \textbf{42.2} & \textbf{33.5} & \textbf{5.1} & \textbf{48.3} & \textbf{42.4} & \textbf{41.8} & \textbf{31.6} & \textbf{36.7}\\

        \bottomrule
    \end{tabular}
    }
    \caption{3D instance segmentation results for the \textit{material} aspect on our OpenScan benchmark.}
\label{openscan_compar_material}
\end{table*}

\begin{table*}[h]

\setlength{\tabcolsep}{2pt}

\renewcommand\arraystretch{1.1}
    \resizebox{0.98\textwidth}{!}
    {
    \begin{tabular}{l|ccccccc|c}
        \toprule
     \multirow{2}{*}{\textbf{Method}} & \textbf{Affordance} &  \textbf{Property} & \textbf{Type} & \textbf{Manner} & \textbf{Synonym} & \textbf{Requirement} & \textbf{Element} & \textbf{Mean} \\
    & UpB \textcolor{Red}{\ding{55}} / \textcolor{Green}{\ding{51}}
     & UpB \textcolor{Red}{\ding{55}} / \textcolor{Green}{\ding{51}}
      & UpB \textcolor{Red}{\ding{55}} / \textcolor{Green}{\ding{51}}
       & UpB \textcolor{Red}{\ding{55}} / \textcolor{Green}{\ding{51}}
        & UpB \textcolor{Red}{\ding{55}} / \textcolor{Green}{\ding{51}}
         & UpB \textcolor{Red}{\ding{55}} / \textcolor{Green}{\ding{51}}
          & UpB \textcolor{Red}{\ding{55}} / \textcolor{Green}{\ding{51}}
           & UpB \textcolor{Red}{\ding{55}} / \textcolor{Green}{\ding{51}}
\\
        \midrule
        \multicolumn{9}{c}{\textbf{AP}} \\
        \midrule
        OpenMask3D & 7.2 / 19.3$\,_\text{\textcolor{darkgray}{+12.1}}$ & 7.5 / 26.7$\,_\text{\textcolor{darkgray}{+19.2}}$ & 8.5 / 18.5$\,_\text{\textcolor{darkgray}{+10.0}}$ & 12.8 / 27.7$\,_\text{\textcolor{darkgray}{+14.9}}$ & 16.9 / 29.4$\,_\text{\textcolor{darkgray}{+12.5}}$ & 13.0 / 26.5$\,_\text{\textcolor{darkgray}{+13.5}}$ & 12.2 / 22.1$\,_\text{\textcolor{darkgray}{+9.9}}$ & 9.7 / 21.6$\,_\text{\textcolor{darkgray}{+11.9}}$ \\

        SAI3D & 5.3 / 14.6$\,_\text{\textcolor{darkgray}{+9.3}}$ & 5.8 / 17.7$\,_\text{\textcolor{darkgray}{+11.9}}$ & 7.8 / 17.4$\,_\text{\textcolor{darkgray}{+9.6}}$ & 11.3 / 18.9$\,_\text{\textcolor{darkgray}{+7.6}}$ & 10.0 / 19.3$\,_\text{\textcolor{darkgray}{+9.3}}$ & 10.0 / 19.2$\,_\text{\textcolor{darkgray}{+9.2}}$ & 8.7 / 16.8$\,_\text{\textcolor{darkgray}{+8.1}}$ & 7.6 / 16.8$\,_\text{\textcolor{darkgray}{+9.2}}$ \\
        
        MaskClustering & 6.2 / 15.4$\,_\text{\textcolor{darkgray}{+9.2}}$ & 7.0 / 21.1$\,_\text{\textcolor{darkgray}{+14.1}}$ & 7.1 / 13.9$\,_\text{\textcolor{darkgray}{+6.8}}$ & 11.1 / 18.0$\,_\text{\textcolor{darkgray}{+6.9}}$ & 16.2 / 18.0$\,_\text{\textcolor{darkgray}{+1.8}}$ & 11.3 / 26.3$\,_\text{\textcolor{darkgray}{+15.0}}$& 7.4 / 17.6$\,_\text{\textcolor{darkgray}{+10.2}}$ & 7.9 / 16.9$\,_\text{\textcolor{darkgray}{+9.0}}$ \\
        
        Open3DIS & \textbf{11.9} / \textbf{23.9}$\,_\text{\textcolor{darkgray}{+12.0}}$ & \textbf{12.8} / \textbf{36.0}$\,_\text{\textcolor{darkgray}{+23.2}}$ & \textbf{14.2} / \textbf{24.5}$\,_\text{\textcolor{darkgray}{+10.3}}$ & \textbf{19.2} / \textbf{34.9}$\,_\text{\textcolor{darkgray}{+15.7}}$ & \textbf{26.7} / \textbf{36.1}$\,_\text{\textcolor{darkgray}{+9.4}}$ & \textbf{19.2} / \textbf{35.0}$\,_\text{\textcolor{darkgray}{+15.8}}$ & \textbf{18.7} / \textbf{28.0}$\,_\text{\textcolor{darkgray}{+9.3}}$ & \textbf{15.4} / \textbf{27.7}$\,_\text{\textcolor{darkgray}{+12.3}}$ \\
        \midrule
        \multicolumn{9}{c}{\textbf{AP$_{50}$}} \\
        \midrule
        
        OpenMask3D & 9.1 / 24.3$\,_\text{\textcolor{darkgray}{+15.2}}$ & 10.0 / 34.2$\,_\text{\textcolor{darkgray}{+24.2}}$ & 11.2 / 24.3$\,_\text{\textcolor{darkgray}{+13.1}}$ & 15.4 / 35.1$\,_\text{\textcolor{darkgray}{+19.7}}$ & 19.7 / 34.7$\,_\text{\textcolor{darkgray}{+15.0}}$ & 16.0 / 34.4$\,_\text{\textcolor{darkgray}{+18.4}}$ & 15.4 / 27.2$\,_\text{\textcolor{darkgray}{+11.8}}$ & 12.2 / 27.4$\,_\text{\textcolor{darkgray}{+15.2}}$  \\
        
        SAI3D & 8.4 / 22.0$\,_\text{\textcolor{darkgray}{+13.6}}$ & 8.3 / 27.7$\,_\text{\textcolor{darkgray}{+19.4}}$ & 11.4 / 26.0$\,_\text{\textcolor{darkgray}{+14.6}}$ & 15.7 / 28.4$\,_\text{\textcolor{darkgray}{+12.7}}$ & 16.7 / 28.3$\,_\text{\textcolor{darkgray}{+11.6}}$ & 15.3 / 30.1$\,_\text{\textcolor{darkgray}{+14.8}}$ & 13.6 / 25.1$\,_\text{\textcolor{darkgray}{+11.5}}$ & 11.5 / 25.4$\,_\text{\textcolor{darkgray}{+13.9}}$ \\
        
        MaskClustering &10.7 / 28.6$\,_\text{\textcolor{darkgray}{+17.9}}$ & 12.3 / 38.5$\,_\text{\textcolor{darkgray}{+26.2}}$ & 13.3 / 26.8$\,_\text{\textcolor{darkgray}{+13.5}}$ & 18.4 / 33.5$\,_\text{\textcolor{darkgray}{+15.1}}$ & 30.3 / 34.2$\,_\text{\textcolor{darkgray}{+3.9}}$ & 21.8 / \textbf{49.7}$\,_\text{\textcolor{darkgray}{+27.9}}$ & 13.5 / 32.5$\,_\text{\textcolor{darkgray}{+19.0}}$ & 14.4 / 31.6$\,_\text{\textcolor{darkgray}{+17.2}}$ \\
        
        Open3DIS & \textbf{14.8} / \textbf{29.9}$\,_\text{\textcolor{darkgray}{+15.1}}$ & \textbf{16.0} / \textbf{43.8}$\,_\text{\textcolor{darkgray}{+27.8}}$ & \textbf{17.9} / \textbf{30.6}$\,_\text{\textcolor{darkgray}{+12.7}}$ & \textbf{22.3} / \textbf{43.1}$\,_\text{\textcolor{darkgray}{+20.8}}$ & \textbf{30.6} / \textbf{42.6}$\,_\text{\textcolor{darkgray}{+12.0}}$ & \textbf{24.1} / 43.8$\,_\text{\textcolor{darkgray}{+19.7}}$ & \textbf{21.9} / \textbf{33.5}$\,_\text{\textcolor{darkgray}{11.6}}$  & \textbf{18.9} / \textbf{34.1}$\,_\text{\textcolor{darkgray}{+15.2}}$ \\
        \midrule
        \multicolumn{9}{c}{\textbf{AP$_{25}$}} \\
        \midrule
        
        OpenMask3D & 10.4 / 27.6$\,_\text{\textcolor{darkgray}{+17.2}}$ & 11.6 / 37.8$\,_\text{\textcolor{darkgray}{+26.2}}$ & 13.0 / 27.4$\,_\text{\textcolor{darkgray}{+14.4}}$ & 17.4 / 38.6$\,_\text{\textcolor{darkgray}{+21.2}}$ & 20.6 / 37.4$\,_\text{\textcolor{darkgray}{+16.8}}$ & 18.9 / 39.4$\,_\text{\textcolor{darkgray}{+20.5}}$ & 17.1 / 31.2$\,_\text{\textcolor{darkgray}{+14.1}}$ & 13.9 / 30.9$\,_\text{\textcolor{darkgray}{+17.0}}$\\
        
        SAI3D  & 10.5 / 28.3$\,_\text{\textcolor{darkgray}{+17.8}}$ & 10.7 / 35.4$\,_\text{\textcolor{darkgray}{+24.7}}$ & 13.4 / 33.2$\,_\text{\textcolor{darkgray}{+19.8}}$ & 18.2 / 36.0$\,_\text{\textcolor{darkgray}{+17.8}}$ & 20.0 / 32.9$\,_\text{\textcolor{darkgray}{+12.9}}$ & 18.7 / 39.5$\,_\text{\textcolor{darkgray}{+20.8}}$ & 16.0 / 32.4$\,_\text{\textcolor{darkgray}{+16.4}}$ & 13.8 / 32.4$\,_\text{\textcolor{darkgray}{+18.6}}$ \\
        
        MaskClustering & 13.7 / \textbf{37.2}$\,_\text{\textcolor{darkgray}{+23.5}}$ & 15.8 / \textbf{48.8}$\,_\text{\textcolor{darkgray}{+33.0}}$ & 17.7 / \textbf{35.0}$\,_\text{\textcolor{darkgray}{+17.3}}$ & 23.1 / 44.8$\,_\text{\textcolor{darkgray}{+21.7}}$ &  \textbf{36.6} / \textbf{45.0}$\,_\text{\textcolor{darkgray}{+8.4}}$ &  \textbf{28.2} / \textbf{61.6}$\,_\text{\textcolor{darkgray}{+33.4}}$ & 17.2 / \textbf{42.7}$\,_\text{\textcolor{darkgray}{+25.5}}$ & 18.5 / \textbf{41.0}$\,_\text{\textcolor{darkgray}{+22.5}}$ \\
        
        Open3DIS &  \textbf{16.7} / 32.7$\,_\text{\textcolor{darkgray}{+16.0}}$ &  \textbf{16.8} / 47.1$\,_\text{\textcolor{darkgray}{+30.3}}$ &  \textbf{20.2} / 34.5$\,_\text{\textcolor{darkgray}{+14.3}}$ &  \textbf{24.2} / \textbf{46.0}$\,_\text{\textcolor{darkgray}{+21.8}}$ & 33.1 / 44.9$\,_\text{\textcolor{darkgray}{+11.8}}$ & 25.5 / 47.2$\,_\text{\textcolor{darkgray}{+21.7}}$ &  \textbf{24.7} / 38.6$\,_\text{\textcolor{darkgray}{+13.9}}$ &  \textbf{20.9} / 37.6$\,_\text{\textcolor{darkgray}{+16.7}}$ \\
        \toprule
    \end{tabular}
    }
    \caption{3D instance segmentation results with the upper bound on our OpenScan benchmark, where ``UpB'' denotes upper bound.}
\label{supptab:upper_bound}
\end{table*}

\begin{table*}[h]

\setlength{\tabcolsep}{2pt}

\renewcommand\arraystretch{1.1}
    \resizebox{0.98\textwidth}{!}
    {
    \begin{tabular}{l|cccccccc|c}
        \toprule
     \multirow{2}{*}{\textbf{Method}} & \textbf{Affordance} &  \textbf{Property} & \textbf{Type} & \textbf{Manner} & \textbf{Synonym} & \textbf{Requirement} & \textbf{Element} & \textbf{Material} & \textbf{Mean} \\
    & LLM \textcolor{Red}{\ding{55}} / \textcolor{Green}{\ding{51}}
     & LLM \textcolor{Red}{\ding{55}} / \textcolor{Green}{\ding{51}}
      & LLM \textcolor{Red}{\ding{55}} / \textcolor{Green}{\ding{51}}
       & LLM \textcolor{Red}{\ding{55}} / \textcolor{Green}{\ding{51}}
        & LLM \textcolor{Red}{\ding{55}} / \textcolor{Green}{\ding{51}}
         & LLM \textcolor{Red}{\ding{55}} / \textcolor{Green}{\ding{51}}
          & LLM \textcolor{Red}{\ding{55}} / \textcolor{Green}{\ding{51}}
           & LLM \textcolor{Red}{\ding{55}} / \textcolor{Green}{\ding{51}}
            & LLM \textcolor{Red}{\ding{55}} / \textcolor{Green}{\ding{51}}\\
        \midrule
        \multicolumn{10}{c}{\textbf{AP}} \\
        \midrule
        OpenMask3D & 7.2 / 10.8$\,_\text{\textcolor{darkgray}{+3.6}}$ & 7.5 / 13.9$\,_\text{\textcolor{darkgray}{+6.4}}$ & 8.5 / 8.5$\,_\text{\textcolor{darkgray}{+0}}$ & 12.8 / 14.1$\,_\text{\textcolor{darkgray}{+1.3}}$ & 16.9 / 25.7$\,_\text{\textcolor{darkgray}{+8.8}}$ & 13.0 / 13.4$\,_\text{\textcolor{darkgray}{+0.4}}$ & 12.2 / 12.1$\,_\text{\textcolor{darkgray}{-0.1}}$ & 18.8 / 10.7$\,_\text{\textcolor{darkgray}{-8.1}}$ & 9.9 / 11.7$\,_\text{\textcolor{darkgray}{+1.8}}$ \\

        SAI3D & 5.3 / 7.2$\,_\text{\textcolor{darkgray}{+1.9}}$ & 5.8 / 9.9$\,_\text{\textcolor{darkgray}{+4.1}}$ & 7.8 / 7.5$\,_\text{\textcolor{darkgray}{-0.3}}$ & 11.3 / 14.5$\,_\text{\textcolor{darkgray}{+3.2}}$ & 10.0 / 16.1$\,_\text{\textcolor{darkgray}{+6.1}}$ & 10.0 / 9.3$\,_\text{\textcolor{darkgray}{-0.7}}$ & 8.7 / 7.9$\,_\text{\textcolor{darkgray}{-0.8}}$ & 11.3 / 6.6 $\,_\text{\textcolor{darkgray}{-4.7}}$& 7.7 / 8.6$\,_\text{\textcolor{darkgray}{+0.9}}$ \\
        MaskClustering & 6.2 / 8.2$\,_\text{\textcolor{darkgray}{+2.0}}$ & 7.0 / 8.5$\,_\text{\textcolor{darkgray}{+1.5}}$ & 7.1 / 7.9$\,_\text{\textcolor{darkgray}{+0.8}}$ & 11.1 / 9.1$\,_\text{\textcolor{darkgray}{-2.0}}$ & 16.2 / 11.1$\,_\text{\textcolor{darkgray}{-5.1}}$ & 11.3 / 16.1$\,_\text{\textcolor{darkgray}{+4.8}}$& 7.4 / 11.6$\,_\text{\textcolor{darkgray}{+4.2}}$ & 12.1 / 8.9$\,_\text{\textcolor{darkgray}{-3.2}}$ & 8.1 / 9.5$\,_\text{\textcolor{darkgray}{+1.4}}$ \\
        Open3DIS & \textbf{11.9} / \textbf{15.1}$\,_\text{\textcolor{darkgray}{+3.2}}$ & \textbf{12.8} / \textbf{22.7}$\,_\text{\textcolor{darkgray}{+9.9}}$ & \textbf{14.2} / \textbf{15.4}$\,_\text{\textcolor{darkgray}{+1.2}}$ & \textbf{19.2} / \textbf{21.6}$\,_\text{\textcolor{darkgray}{+2.4}}$ & \textbf{26.7} / \textbf{31.0}$\,_\text{\textcolor{darkgray}{+4.3}}$ & \textbf{19.2} / \textbf{21.3}$\,_\text{\textcolor{darkgray}{+2.1}}$ & \textbf{18.7} / \textbf{17.6}$\,_\text{\textcolor{darkgray}{-1.1}}$ & \textbf{28.3} / \textbf{18.7}$\,_\text{\textcolor{darkgray}{-9.6}}$ & \textbf{15.8} / \textbf{17.8}$\,_\text{\textcolor{darkgray}{+2.0}}$ \\
        \midrule
        \multicolumn{10}{c}{\textbf{AP$_{50}$}} \\
        \midrule
        OpenMask3D & 9.1 / 13.5$\,_\text{\textcolor{darkgray}{+4.4}}$ & 10.0 / 18.7$\,_\text{\textcolor{darkgray}{+8.7}}$ & 11.2 / 11.0$\,_\text{\textcolor{darkgray}{-0.2}}$ & 15.4 / 17.9$\,_\text{\textcolor{darkgray}{+2.5}}$ & 19.7 / 30.7$\,_\text{\textcolor{darkgray}{+11.0}}$ & 16.0 / 17.3$\,_\text{\textcolor{darkgray}{+1.3}}$ & 15.4 / 15.0$\,_\text{\textcolor{darkgray}{-0.4}}$ & 22.1 / 12.6$\,_\text{\textcolor{darkgray}{-9.5}}$ & 12.5 / 14.7$\,_\text{\textcolor{darkgray}{+2.2}}$  \\
        SAI3D & 8.4 / 10.4$\,_\text{\textcolor{darkgray}{+2.0}}$ & 8.3 / 15.7$\,_\text{\textcolor{darkgray}{+7.4}}$ & 11.4 / 10.9$\,_\text{\textcolor{darkgray}{-0.5}}$ & 15.7 / 20.7$\,_\text{\textcolor{darkgray}{+5.0}}$ & 16.7 / 24.0$\,_\text{\textcolor{darkgray}{+7.3}}$ & 15.3 / 15.1$\,_\text{\textcolor{darkgray}{-0.2}}$ & 13.6 / 12.6$\,_\text{\textcolor{darkgray}{-1.0}}$ & 17.1 / 10.0$\,_\text{\textcolor{darkgray}{-7.1}}$ & 11.6 / 12.8$\,_\text{\textcolor{darkgray}{+1.2}}$ \\
        MaskClustering &10.7 / 14.3$\,_\text{\textcolor{darkgray}{+3.6}}$ & 12.3 / 16.3$\,_\text{\textcolor{darkgray}{+4.0}}$ & 13.3 / 15.1$\,_\text{\textcolor{darkgray}{+1.8}}$ & 18.4 / 16.1$\,_\text{\textcolor{darkgray}{-2.3}}$ & 30.3 / 21.9$\,_\text{\textcolor{darkgray}{-8.4}}$ & 21.8 / \textbf{30.5}$\,_\text{\textcolor{darkgray}{+8.7}}$ & 13.5 / \textbf{21.2}$\,_\text{\textcolor{darkgray}{+7.7}}$ & 20.6 / 15.7$\,_\text{\textcolor{darkgray}{-4.9}}$ & 14.6 / 17.5$\,_\text{\textcolor{darkgray}{+2.9}}$ \\
        Open3DIS & \textbf{14.8} / \textbf{18.7}$\,_\text{\textcolor{darkgray}{+3.9}}$ & \textbf{16.0} / \textbf{28.6}$\,_\text{\textcolor{darkgray}{+12.6}}$ & \textbf{17.9} / \textbf{18.8}$\,_\text{\textcolor{darkgray}{+0.9}}$ & \textbf{22.3} / \textbf{26.2}$\,_\text{\textcolor{darkgray}{+3.9}}$ & \textbf{30.6} / \textbf{36.9}$\,_\text{\textcolor{darkgray}{+6.3}}$ & \textbf{24.1} / 26.7$\,_\text{\textcolor{darkgray}{+2.6}}$ & \textbf{21.9} / 21.0$\,_\text{\textcolor{darkgray}{-0.9}}$ & \textbf{33.6} / \textbf{21.8}$\,_\text{\textcolor{darkgray}{-11.8}}$ & \textbf{19.3} / \textbf{21.7}$\,_\text{\textcolor{darkgray}{+2.4}}$ \\
        \midrule
        \multicolumn{10}{c}{\textbf{AP$_{25}$}} \\
        \midrule
        OpenMask3D & 10.4 / 15.3$\,_\text{\textcolor{darkgray}{+4.9}}$ & 11.6 / 21.1$\,_\text{\textcolor{darkgray}{+9.5}}$ & 13.0 / 14.3$\,_\text{\textcolor{darkgray}{+1.3}}$ & 17.4 / 19.9$\,_\text{\textcolor{darkgray}{+2.5}}$ & 20.6 / 33.2$\,_\text{\textcolor{darkgray}{+12.6}}$ & 18.9 / 20.3$\,_\text{\textcolor{darkgray}{+1.4}}$ & 17.1 / 17.0$\,_\text{\textcolor{darkgray}{-0.1}}$ & 25.0 / 14.1$\,_\text{\textcolor{darkgray}{-10.9}}$ & 14.2 / 17.1$\,_\text{\textcolor{darkgray}{+2.9}}$\\
        SAI3D  & 10.5 / 13.2$\,_\text{\textcolor{darkgray}{+2.7}}$ & 10.7 / 20.2$\,_\text{\textcolor{darkgray}{+9.5}}$ & 13.4 / 14.6$\,_\text{\textcolor{darkgray}{+1.2}}$ & 18.2 / 23.3$\,_\text{\textcolor{darkgray}{+5.1}}$ & 20.0 / 28.3$\,_\text{\textcolor{darkgray}{+8.3}}$ & 18.7 / 18.9$\,_\text{\textcolor{darkgray}{+0.2}}$ & 16.0 / 15.8$\,_\text{\textcolor{darkgray}{-0.2}}$ & 22.9 / 13.4$\,_\text{\textcolor{darkgray}{-9.5}}$ & 14.1 / 16.2$\,_\text{\textcolor{darkgray}{+2.1}}$ \\
        MaskClustering & 13.7 / 17.7$\,_\text{\textcolor{darkgray}{+4.0}}$ & 15.8 / 20.4$\,_\text{\textcolor{darkgray}{+4.6}}$ & 17.7 / 18.5$\,_\text{\textcolor{darkgray}{+0.8}}$ & 23.1 / 22.4$\,_\text{\textcolor{darkgray}{-0.7}}$ &  \textbf{36.6} / 26.7$\,_\text{\textcolor{darkgray}{-9.9}}$ &  \textbf{28.2} / \textbf{36.0}$\,_\text{\textcolor{darkgray}{+7.8}}$ & 17.2 / \textbf{25.5}$\,_\text{\textcolor{darkgray}{+8.3}}$ & 25.6 / 19.9$\,_\text{\textcolor{darkgray}{-5.7}}$ & 18.7 / 21.5$\,_\text{\textcolor{darkgray}{+2.8}}$ \\
        Open3DIS &  \textbf{16.7} / \textbf{20.4}$\,_\text{\textcolor{darkgray}{+3.7}}$ &  \textbf{16.8} / \textbf{30.2}$\,_\text{\textcolor{darkgray}{+13.4}}$ &  \textbf{20.2} / \textbf{20.4}$\,_\text{\textcolor{darkgray}{+0.2}}$ &  \textbf{24.2} / \textbf{28.2}$\,_\text{\textcolor{darkgray}{+4.0}}$ & 33.1 / \textbf{39.2}$\,_\text{\textcolor{darkgray}{+6.1}}$ & 25.5 / 28.5$\,_\text{\textcolor{darkgray}{+3.0}}$ &  \textbf{24.7} / 23.4$\,_\text{\textcolor{darkgray}{-1.3}}$ &  \textbf{36.7} / \textbf{24.3}$\,_\text{\textcolor{darkgray}{-12.4}}$ &  \textbf{21.4} / \textbf{23.6}$\,_\text{\textcolor{darkgray}{+2.2}}$ \\
        \toprule
    \end{tabular}
    }
    \vspace{1mm}
    \caption{3D instance segmentation results with LLM for attribute understanding on our OpenScan benchmark.} 
\label{supptab:llm_result}
\end{table*}

\subsection{Results Without Query Templates.}

In this paper, we adopt query templates (\eg, ``\textit{this term is made of wood}'') as the default experimental configuration. For template-free evaluation on the GOV-3D benchmark (\eg, ``\textit{wood}''), detailed results are provided in Table~\ref{main_result_w/o_template}. We evaluate OpenMask3D~\cite{takmaz2023openmask3d}, SAI3D~\cite{sai3d}, MaskClustering~\cite{yan2024maskclustering}, and Open3DIS~\cite{nguyen2023open3dis} for 3D instance segmentation. Our experiments show that Open3DIS achieves the highest AP, AP$_{50}$, and AP$_{25}$ scores across every linguistic aspect. This performance aligns with its strong performance in the GOV-3D task with query templates.

\subsection{Results of Visual Attributes}

We present comparative visual attribute results for the \textit{material} aspect on our OpenScan benchmark in Table~\ref{openscan_compar_material}. Our evaluation involves OpenMask3D~\cite{takmaz2023openmask3d}, SAI3D~\cite{sai3d}, MaskClustering~\cite{yan2024maskclustering}, and Open3DIS~\cite{nguyen2023open3dis} across 10 \textit{material} attributes. It demonstrates that these OV-3D models perform strongly on the ``porcelain'' material, indicating that the visual information of the ``porcelain'' material in 3D objects (\eg, ``toilet'' and ``bathtub'') is more distinguishable than that of other materials. However, these OV-3D models struggle to accurately segment the ``stone'' material. This difficulty stems from the fact that stone is commonly associated with large 3D regions (\eg, ``wall'' and ``floor''), which are often neglected following the common practice~\cite{mask3d, takmaz2023openmask3d, sai3d} during 3D segmentation. These OV-3D models cannot correctly segment these large 3D areas, resulting in low results of the ``stone'' material. Notably, Open3DIS shows impressive results on each material compared to other OV-3D models, aligning with its strong performance in the classic OV-3D task.

\subsection{Results of Upper Bound}

During the 3D prediction in the GOV-3D task, we query the attributes to obtain the attribute-related 3D mask predictions. For annotating the OpenScan benchmark, object classes are associated with corresponding attributes using the ConceptNet~\cite{speer2017conceptnet} database. Conversely, each attribute query can also be associated with the corresponding object classes. Therefore, we can replace the attribute queries with the ground truth attribute-related object classes from ConceptNet to finalize 3D mask results. We exclude the \textit{material} aspect since the related object classes of \textit{material} in the ConceptNet database are limited. We serve this setting as our upper bound performance. Table~\ref{supptab:upper_bound} shows the comparison of baseline methods OpenMask3D~\cite{takmaz2023openmask3d}, SAI3D~\cite{sai3d}, MaskClustering~\cite{yan2024maskclustering}, and Open3DIS~\cite{nguyen2023open3dis} with their upper bounds, highlighting significant performance gaps that underscore the potential for attribute-aware 3D reasoning.

\subsection{Results of Introducing LLM for Attribute Understanding}

In our failure case analysis in the main paper, we observe that the OV-3D model Open3DIS~\cite{nguyen2023open3dis} can identify the object classes (\eg, ``piano'') in the OV-3D task but fails to recognize the associated object attributes (\eg, ``this term has 88 keys'') in the GOV-3D task. This suggests a promising direction for improving GOV-3D performance by leveraging large language models (LLMs) to perform high-level reasoning, transforming the GOV-3D attribute queries (\eg, ``this term has 88 keys'') back to the OV-3D class queries (\eg, ``piano'').
To this end, we design an experiment where an LLM (\ie, Vicuna-7B~\cite{vicuna2023}) is prompted to map object attributes to corresponding object classes. Given an attribute [attribute], the LLM is prompted as:

\begin{center}
\textit{Q: Given an object's attribute [attribute], please output the related object's classes in the indoor scene separated by commas.}
\end{center}

The LLM will generate a list of object classes corresponding to the input attribute [attribute]. We then format the object classes into a sentence as a query for evaluation. We utilize the baseline methods OpenMask3D~\cite{takmaz2023openmask3d}, SAI3D~\cite{sai3d}, MaskClustering~\cite{yan2024maskclustering}, and Open3DIS~\cite{nguyen2023open3dis} to compare their performance on whether introducing LLM for attribute understanding. As shown in Table~\ref{supptab:llm_result}, introducing LLM can improve the attribute understanding performance across most linguistic aspects of the GOV-3D task. However, performance declines in the \textit{material} aspect, as LLMs rely exclusively on linguistic inputs and lack visual context required to differentiate material properties (\eg ``wooden chair'' and ``plastic chair''). Additionally, the LLM’s output can be noisy and inconsistent, occasionally producing object classes unrelated to the input attribute, which degrades performance in some linguist aspects.

\subsection{Results of Weighted Mean}
\vspace{1mm}

In the OpenScan benchmark, we observe disparities in the attribute annotations for linguistic aspects (\eg, ``affordance'' and ``synonym''). To address the imbalance in attribute annotations, we introduce a weighted mean score (w-Mean) metric that normalizes contributions based on annotation counts. For linguist aspects with annotation counts $L=\{l_k\}_{k=1}^{H}$ and corresponding scores $S=\{s_k\}_{k=1}^{H}$, the w-Mean is computed as:

\begin{equation}
\text{w-Mean} = \frac{\sum_{k=1}^{H} s_{k}l_{k}}{\sum_{k=1}^{H}l_{k}}
\end{equation}

\vspace{1mm}

As shown in Table~\ref{suppl:w-mean}, applying the w-Mean metric improves the performance of baseline methods, including OpenMask3D~\cite{takmaz2023openmask3d}, SAI3D~\cite{sai3d}, MaskClustering~\cite{yan2024maskclustering}, and Open3DIS~\cite{nguyen2023open3dis}. This improvement steams from the w-Mean metric’s ability to normalize the contribution of each linguistic aspect based on its annotation count, thus mitigating biases from attributes and enhancing the robustness of performance evaluation across linguistic aspects in the GOV-3D task.

\begin{table}[t]
\centering
    \resizebox{0.46\textwidth}{!}
    {
    \begin{tabular}{l|ccc|ccc}
        \toprule
      \multirow{2}{*}{\textbf{Method}} & \multicolumn{3}{c|}{\textbf{Mean}} & \multicolumn{3}{c}{\textbf{w-Mean}} \\
      
      & \textbf{AP} & \textbf{AP$_{50}$} & \textbf{AP$_{25}$} & \textbf{AP} & \textbf{AP$_{50}$} & \textbf{AP$_{25}$} \\
      \midrule
       OpenMask3D       & 9.9  & 12.5 & 14.2 & 12.3 & 15.0 & 17.1\\
       SAI3D            & 7.7  & 11.6 & 14.1 & 8.6 & 13.0 & 16.4\\
       MaskClustering   & 8.1  & 14.6 & 18.7 & 9.0 & 16.0 & 20.3\\
       Open3DIS         & \textbf{15.8} & \textbf{19.3} & \textbf{21.4} & \textbf{19.1} & \textbf{23.1} & \textbf{25.5}\\
        
        \bottomrule
    \end{tabular}
    }
    \vspace{1mm}
    \caption{3D instance segmentation results for weighted-mean score (w-Mean) on our OpenScan benchmark.}
    \vspace{1mm}
\label{suppl:w-mean}
\end{table}

\vspace{4mm}

\begin{figure*}[ht] \centering
    \includegraphics[width=0.98\textwidth]{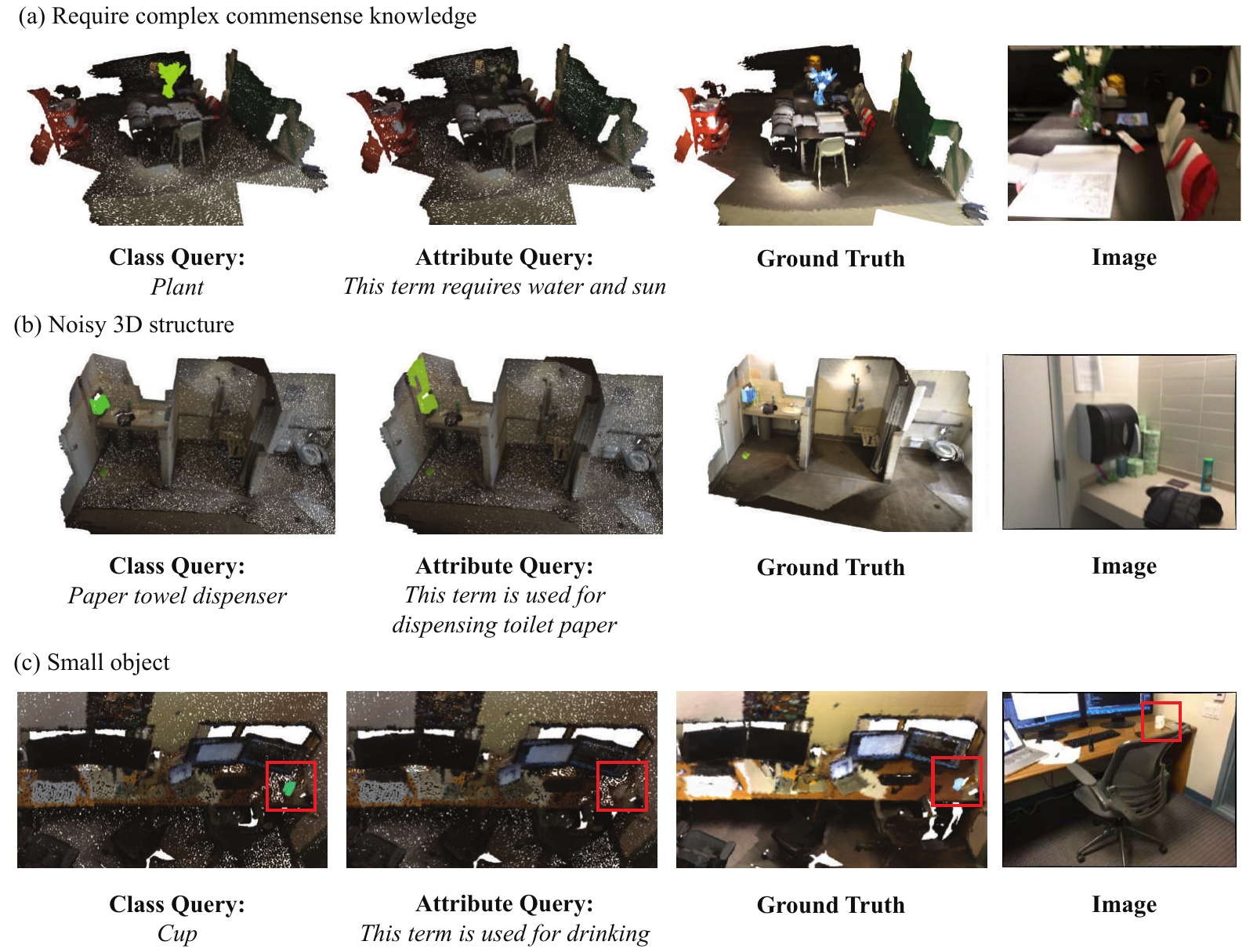}
    \vspace{10mm}
    \caption{\zyjn{Visualization of the Open3DIS failure cases. The ground truth objects and outputs are highlighted in color. Best view with zoom in.}}
    \label{failure_case_supp}
    \vspace{4mm}
\end{figure*}

\begin{figure*}[ht]
    \centering
    \includegraphics[width=0.98\textwidth]{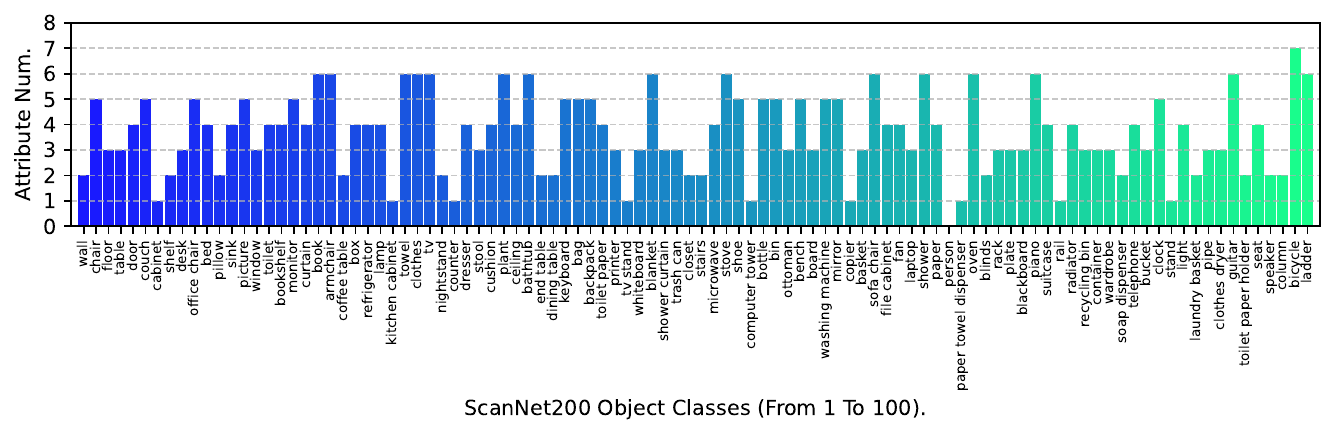}
    \includegraphics[width=0.98\textwidth]{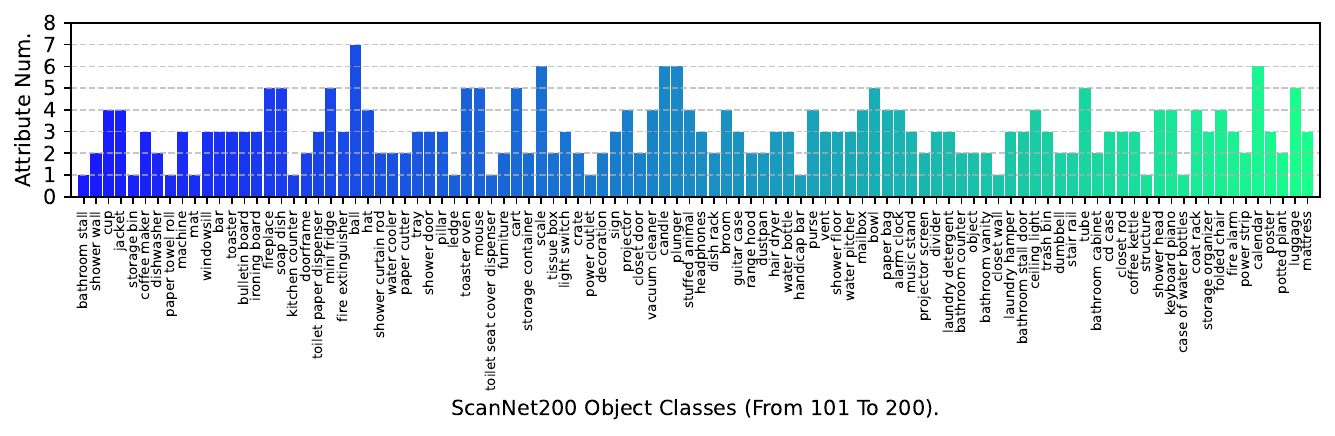}
    \caption{Number of attributes per object class from ScanNet200 in our OpenScan benchmark.}
    \label{attributes_num_per_object_class}
\end{figure*}

\begin{figure}[h] \centering
    \includegraphics[width=0.46\textwidth]{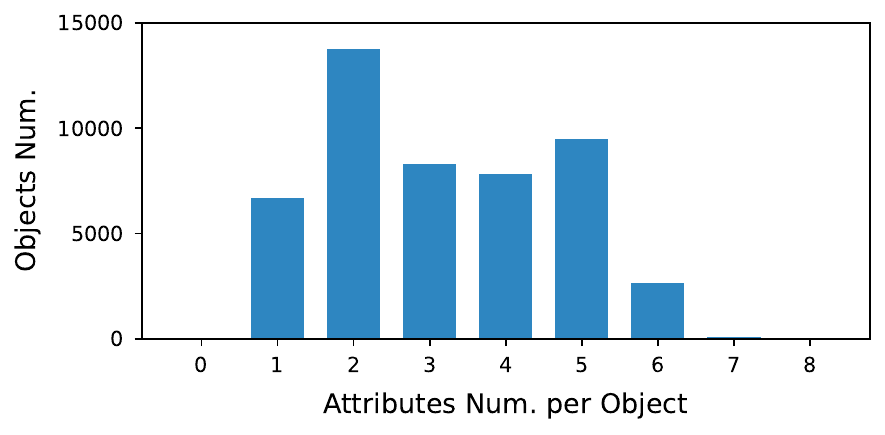}
    \caption{Number of attributes per object in our OpenScan benchmark and corresponding number of objects.}

    \label{attribute_per_object}
\end{figure}

\begin{figure}[h] \centering
    \includegraphics[width=0.46\textwidth]{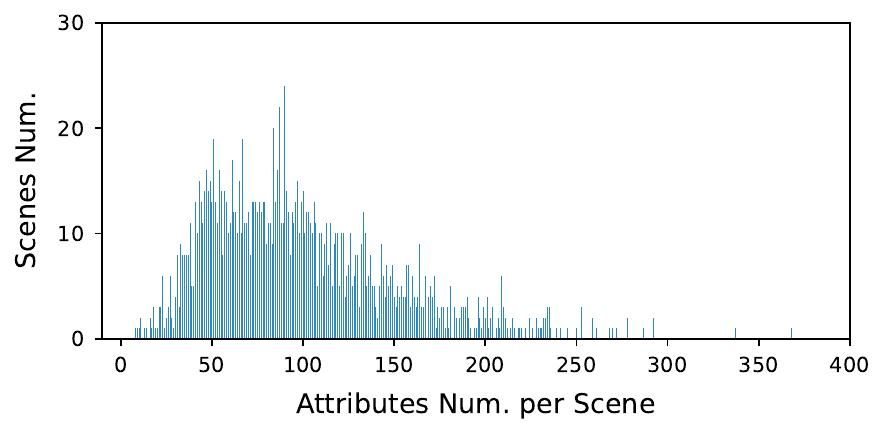}
    \caption{Number of attributes per scene in our OpenScan benchmark and corresponding number of scenes.}
    \label{attribute_per_scene}
\end{figure}

\begin{figure*}[ht] \centering
    \includegraphics[width=0.95\textwidth]{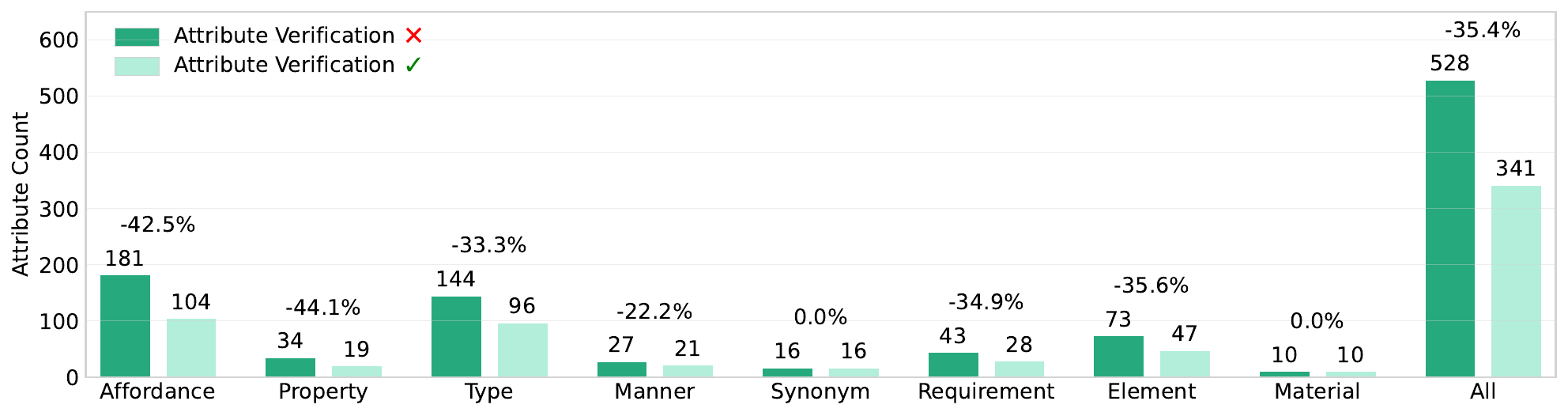}
    \vspace{2mm}
    \caption{OpenScan benchmark statistics of attributes during attribute verification.}
    \label{attribute_verification}
    \vspace{2mm}
\end{figure*}

\begin{figure*}[h]
    \centering 
    \includegraphics[width=0.98\textwidth]{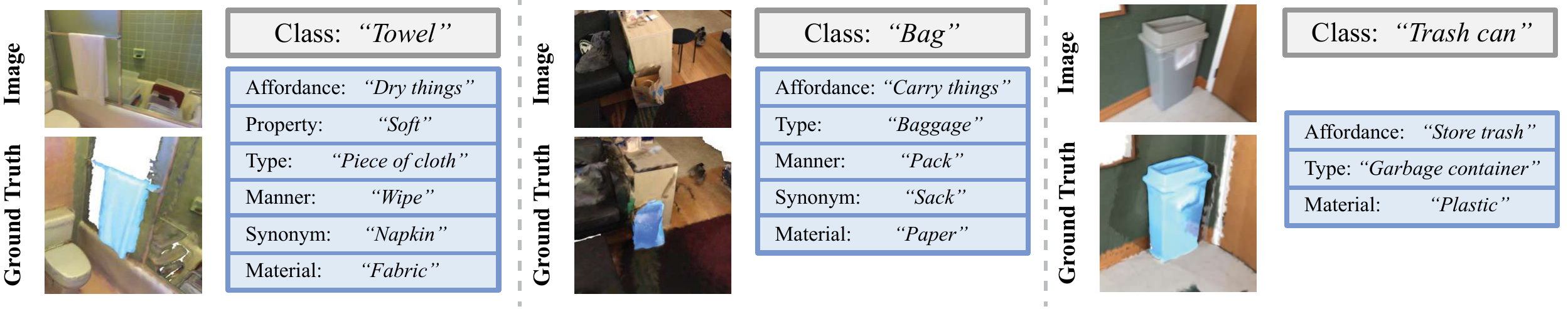}
    \vspace{2mm}
    \caption{Examples of objects and corresponding attributes in our OpenScan benchmark.}
    \label{object_attribute}
    \vspace{2mm}
\end{figure*}

\begin{figure}[h] \centering
    \includegraphics[width=0.46\textwidth]{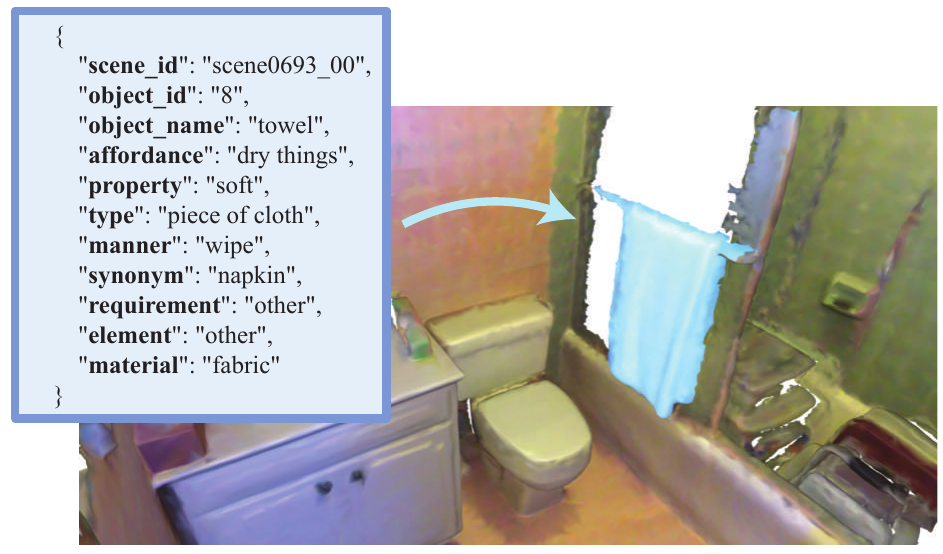}
    \vspace{2mm}
    \caption{OpenScan benchmark format. The target object is highlighted in blue.}
    \label{format}
\end{figure}

\subsection{Additional Failure Cases Analysis}

\zyja{As illustrated in Figure~\ref{failure_case_supp}, the Open3DIS model\cite{nguyen2023open3dis} for OV-3D exhibits limitations in the GOV-3D task under specific conditions. Specifically, the model struggles to generate accurate 3D masks when: (a) the attribute query requires complex commonsense knowledge (\eg, ``this term requires water and sun''), resulting in failure to predict 3D masks; (b) the target 3D object contains noisy geometry, such as 3D holes or irregular 3D structures, leading to partially incorrect 3D masks; and (c) the target object is small, providing insufficient geometric detail for segmentation, causing the model to fail in predicting 3D masks. In contrast, Open3DIS correctly predicts 3D masks for attribute-related class queries in the OV-3D task under these scenarios, underscoring the challenge of the GOV-3D task.}

\section{Additional Benchmark Details}
\label{add_benchmark_detail}
\vspace{1mm}

\subsection{Does OpenScan Represent 200 Object Classes From ScanNet200 Well Enough?}
\vspace{1mm}

During the annotation of our OpenScan benchmark, object classes from ScanNet200~\cite{scannet200} are labeled with attributes using the ConceptNet~\cite{speer2017conceptnet} database and manual annotation.
Figure~\ref{attributes_num_per_object_class} shows the number of attributes per object class from ScanNet200 in our OpenScan benchmark. Notably, most object classes from ScanNet200 are annotated with more than one attribute in our OpenScan, indicating that our OpenScan benchmark adequately represents object classes from ScanNet200. Besides, the object class has up to seven attributes (\ie, ``\textit{bicycle}'', and ``\textit{ball}'') in our OpenScan benchmark.

\subsection{Number of Attributes per Object.}
\vspace{1mm}

Figure~\ref{attribute_per_object} summarizes the distribution of attributes per object in our OpenScan benchmark. The majority of objects have 1–6 attributes.

\subsection{Number of Attributes per Scene.}
Figure~\ref{attribute_per_scene} presents the distribution of attributes per scene in our OpenScan benchmark. It demonstrates that the attributes in 3D scenes are semantically rich.

\subsection{Attribute Verification in Benchmark Annotation}

\zyja{During the attribute annotation process, we leverage a knowledge graph to automatically generate object-related attributes. However, the initial attribute set often contains noise, including attributes that are irrelevant, ambiguous, or semantically inconsistent with the related object classes. To address this, we conduct a meticulous manual verification process to refine the attribute set, ensuring semantic consistency and coherence in our OpenScan benchmark.}

\zyja{As shown in Figure~\ref{attribute_verification}, our OpenScan benchmark generates 528 attributes initially. After attribute verification, 341 attributes are retained, resulting in an overall reduction of 35.4\% of noisy attributes. Notably, the \textit{affordance} aspect exhibits high noise level, with 42.5\% of attributes being filtered out, suggesting that the \textit{affordance} attributes are particularly prone to ambiguity due to the diverse nature of affordance candidates within the knowledge graph.
In contrast, all \textit{synonym} attributes are retained during verification. This robustness is attributed to the high semantic similarity between synonym attributes and their corresponding object classes, ensuring reliable alignment during the generation process.}

\subsection{Additional Benchmark Samples}
\begin{figure*}[h] \centering
    \includegraphics[width=0.79\textwidth]{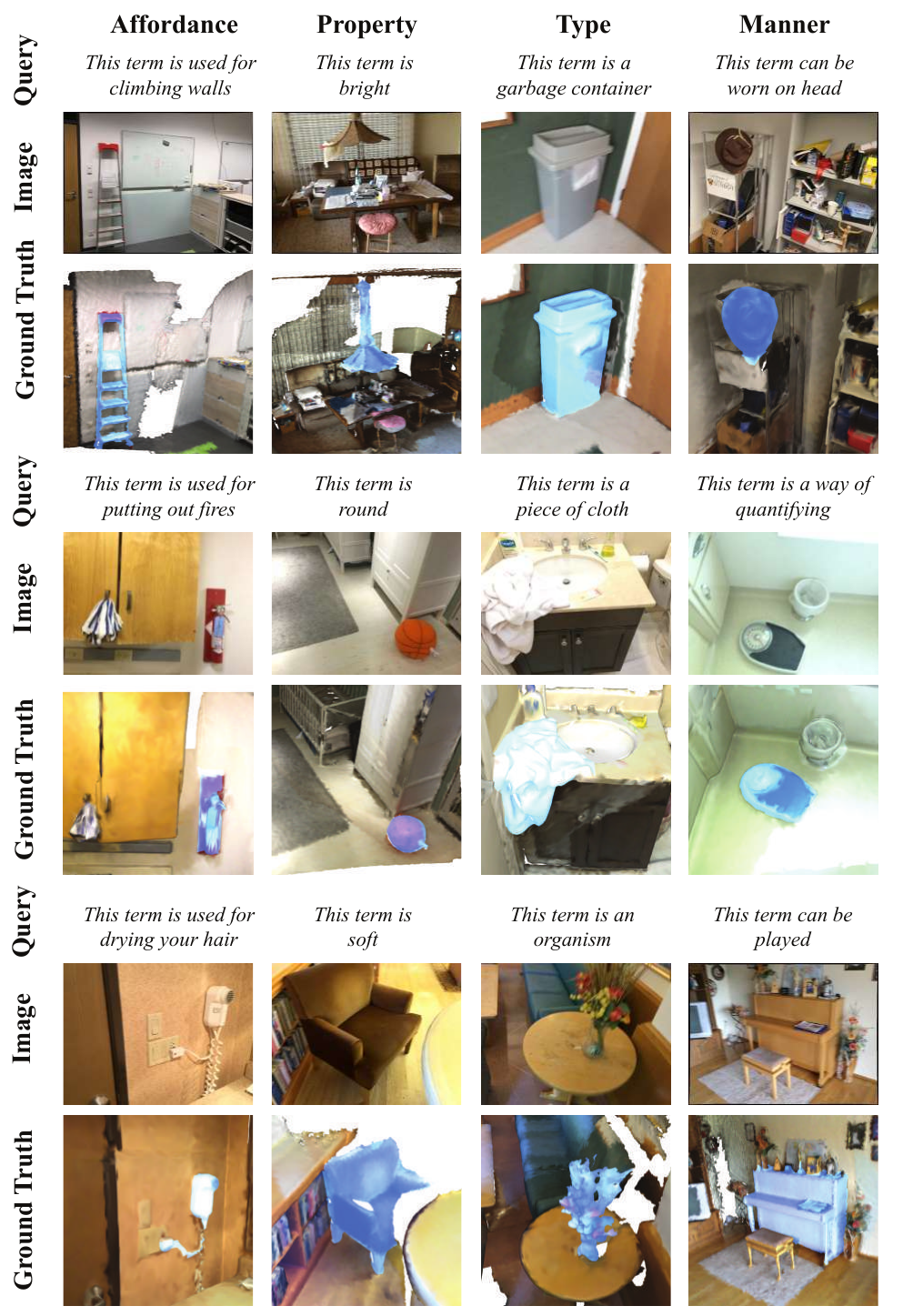}
    \caption{Additional OpenScan benchmark samples of \textit{affordance}, \textit{property}, \textit{type}, and \textit{manner} aspects. Target objects are highlighted in blue.}
    \label{dataset_sample_supp_1}
    \vspace{-4mm}
\end{figure*}

\begin{figure*}[h] \centering
    \includegraphics[width=0.79\textwidth]{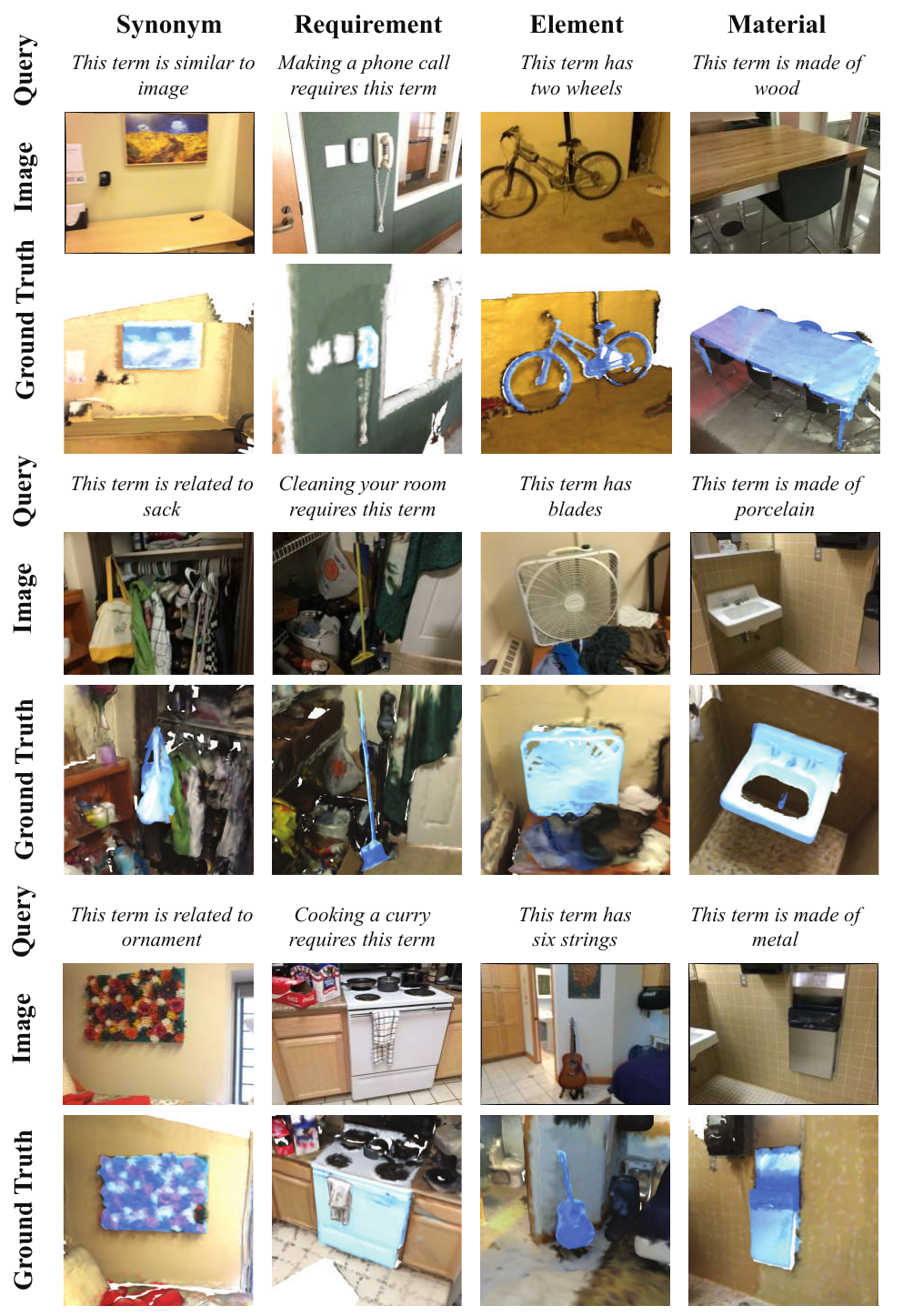}
    \caption{Additional OpenScan benchmark samples of \textit{synonym}, \textit{requirement}, \textit{element}, and \textit{material} aspects. Target objects are highlighted in blue.}
    \label{dataset_sample_supp_2}
    \vspace{-2mm}
\end{figure*}

We provide additional samples of our OpenScan benchmark. Figure~\ref{object_attribute} presents the examples of objects and their corresponding attributes. Figure~\ref{dataset_sample_supp_1} displays the \textit{affordance}, \textit{property}, \textit{type}, and \textit{manner} aspects, while Figure~\ref{dataset_sample_supp_2} shows the \textit{synonym}, \textit{requirement}, \textit{element}, and \textit{material} aspects.

\subsection{Benchmark Formats}

Figure~\ref{format} shows an example of our OpenScan benchmark formats. Our OpenScan is formatted in the JSON file. Each target 3D object is annotated with the following items:

\begin{itemize}
\item \textbf{\textit{Scene ID}}: indicates the scene in which the target object is located.

\item \textbf{\textit{Object ID}}: identifies the target object's unique ID within the scene.

\item \textbf{\textit{Object Name}}: specifies the object class of the target object.
\end{itemize}

In addition, each object is annotated with eight linguistic aspects (\textit{affordance}, \textit{property}, \textit{type}, \textit{manner}, \textit{synonym}, \textit{requirement}, \textit{element}, and \textit{material}). If the target 3D object does not contain an attribute of a specific linguistic aspect, it is marked as ``other''.

\subsection{Benchmark Details}
We construct our OpenScan benchmark based on ScanNet200~\cite{scannet200} across eight linguistic aspects. We present all attributes and their corresponding query templates in our OpenScan benchmark: Table~\ref{dataset_example_1} displays the \textit{affordance} and \textit{property} aspects; Table~\ref{dataset_example_2} shows the \textit{type}, \textit{manner}, and \textit{synonym} aspects; and Table~\ref{dataset_example_3} presents the \textit{requirement}, \textit{element}, and \textit{material} aspects. The ``object'' in the query template is replaced with ``this term'' in our experiments.

\section{Additional Related Work}
\label{add_related}

\noindent\textbf{Open-Vocabulary 2D Understanding Benchmarks.}
Open-vocabulary 2D understanding refers to the task of detecting or segmenting novel \zyjn{object classes} \rl{that are not present in the training dataset. For the} object detection task, COCO~\cite{coco} and LVIS~\cite{lvis} are two widely used datasets. For \rl{the} image segmentation task, popular datasets include COCO~\cite{coco}, ADE20k~\cite{ade20k}, PASCAL-VOC~\cite{everingham2015pascal}, and Cityscapes~\cite{cordts2016cityscapes}. However, these benchmarks primarily evaluate the model's open-vocabulary ability but do not explicitly assess its capability to recognize specific object characteristics. PACO~\cite{ramanathan2023paco} introduces a 2D segmentation benchmark that focuses on parts and attributes of common objects. Inspired by PACO~\cite{ramanathan2023paco}, FG-OVD~\cite{fg-ovd} presents a challeng\zyjn{ing} task and benchmark for fine-grained open-vocabulary object detection to evaluate the ability of open-vocabulary detectors to discern extrinsic object properties. Similarly, OVDEval~\cite{ovdeval} introduces an open-vocabulary detection benchmark to evaluate the performance on linguistic aspects using complex language prompts. Our work is different from them~\cite{ramanathan2023paco,fg-ovd,ovdeval} since we focus on the understanding of object attributes on 3D data, which poses greater challenges compared to \rl{the} understanding in 2D images due to the limited annotations in 3D benchmarks.

\section{Limitations and Future Work}
\label{limit_future}
Our benchmark is currently constructed solely on the ScanNet200 benchmark with limited 3D indoor scene. It would be beneficial to increase the scale of our benchmark to include a wider variety of 3D scenes and objects.
In future work, we plan to extend our OpenScan benchmark to encompass more diverse scenes by incorporating indoor 3D datasets such as ScanNet++~\cite{scannet++} and Matterport3D~\cite{Matterport3D}. Our mature annotation procedures can be readily adapted to these datasets. Moreover, we aim to evaluate current OV-3D models on our GOV-3D task, particularly examining performance variations when using higher point resolutions in ScanNet++~\cite{scannet++} and larger scene areas in Matterport3D~\cite{Matterport3D}.

\section{Broader Impact}
\label{broader_impact}
Our approach does not introduce any negative societal impacts. All experiments are performed on publicly available datasets, with no use of private data. Although our benchmark is constructed exclusively from public data, we recognize the potential for unintended consequences if the data is applied without appropriate safeguards. We urge readers to ensure that the application of this research remains lawful and ethical, strictly adhering to established regulations and guidelines.

\begin{table*}[h]
\centering
    \resizebox{1\textwidth}{!}
    {
    \begin{tabular}{c|cc|cc}
        \toprule
     & \textbf{Attribute} & \textbf{Template} &  \textbf{Attribute} & \textbf{Template}  \\
        \midrule
    \multirow{52}{*}{\rotatebox{90}{\textbf{Affordance}}} 
&  carry things & [object] is used for carrying things 
&  holding up a roof  & [object] is used for holding up a roof \\
&  rest &  [object] is used for resting
&  sit & [object] is used for sitting \\
&  keep food cold  & [object] is used for keeping food cold
&  place coffee & [object] is used for placing coffee \\
&  work & [object] is used for working 
& look outside & [object] is used for looking outside \\
& bath  &  [object] is used for bathing 
& cover a window & [object] is used for covering a window \\
& stand & [object] is used for standing 
& wash dishes & [object] is used for washing dishes \\
& measure weight & [object] is used for measuring weight
& store trash & [object] is used for storing trash \\
& display images  & [object] is used for displaying images
& sleep & [object] is used for sleeping \\
& poop & [object] is used for pooping
& tell time & [object] is used for telling time \\
& bake toaster & [object] is used for baking toaster
& perform music & [object] is used for performing music \\
& making toast & [object] is used for making toast
& heat food & [object] is used for heating food \\
& separate rooms & [object] is used for separating rooms
& close the top of a room & [object] is used for closing the top of a room \\
& ride & [object] is used for riding
& store books & [object] is used for storing books \\
& see yourself & [object] is used for seeing yourself
& store guitar & [object] is used for storing guitar \\
& dry things & [object] is used for drying things
& put your feet on & [object] is used for putting your feet on \\
& storage dirty clothes  &  [object] is used for storaging dirty clothes
& hold up the roof  &  [object] is used for holding up the roof \\
& represent  & [object] is used for representing
& hang clothes & [object] is used for hanging clothes \\
& heat the room  & [object] is used for heating the room
& make coffee & [object] is used for making coffee \\
& presenting information & [object] is used for presenting information
& grow in a garden  &  [object] is used for growing in a garden \\
& cool a person  & [object] is used for cooling a person
& foot protection  & [object] is used for foot protection \\
& heat a room  &  [object] is used for heating a room
& illuminate an area  &  [object] is used for illuminating an area \\
& protecting your head & [object] is used for protecting your head
& print documents &  [object] is used for printing documents \\
& store liquids  &  [object] is used for storing liquids
& keep out light from houses  &  [object] is used for keeping out light from houses \\
& transport things  & [object] is used for transporting things
& collect recyclable plastics  &  [object] is used for collecting recyclable plastics \\
& communicate  & [object] is used for communicating
& pack clothes for a trip  &  [object] is used for packing clothes for a trip \\
& carrying money & [object] is used for carrying money
& wear & [object] is used for wearing \\
& learning  &  [object] is used for learning
& store things  & [object] is used for storing things \\
& carry liquids  &   [object] is used for carrying liquids
& turn on a light  &  [object] is used for turning on a light \\
& write ideas and terms on  & [object] is used for writing ideas and terms on
& store file  & [object] is used for storing file \\
& make a captured voice become audible  &  [object] is used for making a captured voice become audible
& type  & [object] is used for typing \\
& eat dinner & [object] is used for eating dinner
& bake cookies  &  [object] is used for baking cookies \\
& furnish  & [object] is used for furnishing
& detect fire & [object] is used for detecting fire \\
& have privacy  &  [object] is used for having privacy
& hold toilet paper  &  [object] is used for holding toilet paper \\
& blow your nose  & [object] is used for blowing your nose
& store water  &  [object] is used for storing water \\
& bounce  &  [object] is used for bouncing
& cover a bed  &  [object] is used for covering a bed \\
& organize books  & [object] is used for organizing books
& hold trash & [object] is used for holding trash \\
& climb & [object] is used for climbing
& store clothes & [object] is used for storing clothes \\
& drink  &  [object] is used for drinking
& listen to music  &  [object] is used for listening to music \\
& hold sheet music  &  [object] is used for holding sheet music
& unblocking a toilet  &  [object] is used for unblocking a toilet \\
& hang clothes  &  [object] is used for hanging clothes
& entertain a child  &  [object] is used for entertaining a child \\
& control a computer  &  [object] is used for controlling a computer
& dispense toilet paper  &  [object] is used for dispensing toilet paper \\
& keep clothes & [object] is used for keeping clothes
& entry and exit to the shower  &  [object] is used for entry and exit to the shower \\
& climb walls  &  [object] is used for climbing walls
& hold soap & [object] is used for holding soap \\
& hold things  &  [object] is used for holding things
& get drunk  &  [object] is used for getting drunk \\
& putting out fires  &  [object] is used for putting out fires
& carry something  &  [object] is used for carrying something \\
& hang coat  &  [object] is used for hanging coat
& spray water  &  [object] is used for spraying water \\
& hold food  &  [object] is used for holding food
& dry your hair &  [object] is used for drying your hair \\
& show movies & [object] is used for showing movies
& dry clothes  &  [object] is used for drying clothes \\
& wash clothes  &  [object] is used for washing clothes
& mark that special date  &  [object] is used for marking that special date \\
& vacumming  &  [object] is used for vacumming
& ironing clothes  &  [object] is used for ironing clothes \\
& decorating your room & [object] is used for decorating your room
& sweeping & [object] is used for sweeping \\
& receiving letters & [object] is used for receiving letters
& hold cd & [object] is used for holding cd \\

\midrule
\multirow{10}{*}{\rotatebox{90}{\textbf{Property}}}

& useful for camping  &  [object] is useful for camping
& soft  &  [object] is soft \\
& opaque and closed  &  [object] is opaque and closed
& essential for privacy  &  [object] is essential for privacy \\
& helpful in making comparisons  &  [object] is helpful in making comparisons
& analog or digital  &  [object] is analog or digital \\
& hot & [object] is hot
& one kind of stringed instrument & [object] is one kind of stringed instrument \\
& open or closed & [object] is open or closed
& horizontal & [object] is horizontal \\
& fun to ride & [object] is fun to ride
& reflective & [object] is reflective \\
& alive & [object] is alive
& bright & [object] is bright \\
& hollow & [object] is hollow
& round & [object] is round \\
& useful for unblocking a toilet & [object] is useful for unblocking a toilet
& shaped like a shell & [object] is shaped like a shell \\
& convex down & [object] is convex down
& & \\

        \bottomrule
    \end{tabular}
        }
    \caption{OpenScan benchmark attributes of \textit{affordance} and \textit{property} aspects.}
\label{dataset_example_1}

\end{table*}

\begin{table*}[h]
\centering
    \resizebox{1\textwidth}{!}
    {
    \begin{tabular}{c|cc|cc}
        \toprule
     & \textbf{Attribute} & \textbf{Template} &  \textbf{Attribute} & \textbf{Template}  \\

\midrule
\multirow{48}{*}{\rotatebox{90}{\textbf{Type}}} 

& baggage & [object] is a baggage
& seat & [object] is a seat \\
& table were someone works & [object] is a table were someone works
& plumbing fixture & [object] is a plumbing fixture \\
& window covering &  [object] is a window covering
& land & [object] is a land  \\
& measuring instrument & [object] is a measuring instrument
& garbage container & [object] is a garbage container \\
& a way to relax & [object] is a way to relax
& a good place to lie  &  [object] is a good place to lie \\
& vanity & [object] is a vanity
& kitchen appliance & [object] is a kitchen appliance \\
& basket & [object] is a basket
& box & [object] is a box \\
& string instrument & [object] is a string instrument
& rack & [object] is a rack \\
& appliances & [object] is an appliance
& movable barrier & [object] is a movable barrier \\
& upper surface & [object] is an upper surface
& a two wheel vehicle & [object] is a two wheel vehicle \\
& reflector & [object] is a reflector
& container & [object] is a container \\
& piece of cloth & [object] is a piece of cloth
& representation & [object] is a representation \\
& clue & [object] is a clue
& organism & [object] is an organism \\
& a cooling device & [object] is a cooling device
& footwear & [object] is a footwear \\
& heater & [object] is a heater
& source of illumination & [object] is a source of illumination \\
& a form of clothing & [object] is a form of clothing
& refrigerator & [object] is a refrigerator \\
& dispenser & [object] is a dispenser
& a long seat with no backrest & [object] is a long seat with no backrest \\
& a vehicle & [object] is a vehicle
& bin & [object] is a bin \\
& a communication device & [object] is a communication device
& handbag & [object] is a handbag \\
& coat & [object] is a coat
& an excellent source of information & [object] is an excellent source of information \\
& tube & [object] is a tube
& switch & [object] is a switch \\
& sill & [object] is a sill
& door & [object] is a door \\
& board & [object] is a board
& cabinet & [object] is a cabinet \\
& portable computer & [object] is a portable computer
& display & [object] is a display \\
& computer device & [object] is a computer device
& shaft & [object] is a shaft \\
& alarm & [object] is an alarm
& curtain & [object] is a curtain \\
& paper & [object] is a paper
& bottle & [object] is a bottle \\
& an instrument of music & [object] is an instrument of music
& a toy & [object] is a toy \\
& bedclothes & [object] is a bedclothes
& cutting implement & [object] is a cutting implement \\
& shelf & [object] is a shelf
& table & [object] is a table \\
& supporter & [object] is a supporter
& railing & [object] is a railing \\
& trophy & [object] is a trophy
& audio device & [object] is an audio device \\
& vessel & [object] is a vessel
& a tool to unclog toilets & [object] is a tool to unclog toilets \\
& rod & [object] is a rod
& padding & [object] is a padding \\
& bag & [object] is a bag
& toy animal & [object] is a toy animal \\
& a container for clothes & [object] is a container for clothes
& hole & [object] is a hole \\
& stairs & [object] is a stairs
& storage device & [object] is a storage device \\
& firefighting equipment & [object] is a firefighting equipment
& fitness equipment & [object] is a fitness equipment \\
& device for spraying & [object] is a device for spraying
& counter & [object] is a counter \\
& clock & [object] is a clock
& kettle & [object] is a kettle \\
& hood & [object] is a hood
& beauty device & [object] is a beauty device \\
& optical device & [object] is a optical device
& dryer & [object] is a dryer \\
& detergent & [object] is a detergent
& machine & [object] is a machine \\
& screen & [object] is a screen
& drafting instrument & [object] is a drafting instrument \\
& pad & [object] is a pad
& pitcher & [object] is a pitcher \\
& electronic piano & [object] is an electronic piano
& time list & [object] is a time list \\
& household cleaning tool & [object] is a household cleaning tool
& sign & [object] is a sign \\
& receptacle container & [object] is a receptacle container
& a container for letters & [object] is a container for letters \\

\midrule
\multirow{11}{*}{\rotatebox{90}{\textbf{Manner}}} 

& pack & [object] is a way of packing
& observe & [object] is a way of observing \\
& bathe & [object] is a way of bathing
& quantify & [object] is a way of quantifying \\
& cook & [object] is a way of cooking
& steered by handlebars & [object] can be steered by handlebars \\
& wipe & [object] is a way of wiping
& wear & [object] is a way of wearing \\
& worn on a head & [object] can be worn on a head
& transport things & [object] is a way of transporting things \\
& written on & [object] can be written on
& played & [object] can be played \\
& played with & [object] can be played with
& cover bed & [object] is a way of covering bed \\
& used in a toilet & [object] can be used in a toilet
& manipulate computer & [object] is a way of manipulating computer \\
& climbed to reach some place high & [object] can be climbed to reach some place high
& store & [object] is a way of storing \\
& produce & [object] is a way of producing
& lit with a match & [object] can be lit with a match \\
& schedule & [object] is a way of scheduling
& & \\

\midrule
        
\multirow{8}{*}{\rotatebox{90}{\textbf{Synonym}}}

& weight & [object] is related to weight
& news & [object] is related to news \\
& bedside table & [object] is related to bedside table
& napkin & [object] is related to napkin \\
& image & [object] is similar to image
& sack & [object] is related to sack \\
& reading & [object] is related to reading
& pipe & [object] is related to pipe \\
& power bar & [object] is related to power bar
& round & [object] is related to round \\
& ornament & [object] is related to ornament
& suction cup & [object] is similar to suction cup \\
& dress & [object] is related to dress
& houseplant & [object] is related to houseplant \\
& suitcase & [object] is related to suitcase
& almanac & [object] is related to almanac \\

        \bottomrule
    \end{tabular}
        }
    \caption{OpenScan benchmark attributes of \textit{type}, \textit{manner}, and \textit{synonym} aspects.}
\label{dataset_example_2}

\end{table*}

\begin{table*}[h]
\centering
    \resizebox{1\textwidth}{!}
    {
    \begin{tabular}{c|cc|cc}
        \toprule
     & \textbf{Attribute} & \textbf{Template} &  \textbf{Attribute} & \textbf{Template}  \\

\midrule
\multirow{14}{*}{\rotatebox{90}{\textbf{Requirement}}}

& sit down & sitting down requires [object]
& be unpluged & [object] does not desire to be unpluged \\
& have a bath & having a bath requires [object]
& using a VCR & using a VCR requires [object] \\
& wake up in the morning &  waking up in the morning requires [object]
& playing a guitar & playing a guitar requires [object] \\
& balance to ride & [object] requires balance to ride
& grooming & grooming requires [object] \\
& get warm & getting warm requires [object]
& water and sun & [object] requires water and sun \\
& print & printing requires [object] 
& drink & drinking requires [object] \\
& buying food & buying food requires [object]
& make a phone call & making a phone call requires [object] \\
& bring suit & bringing suit requires [object]
& go on the internet & going on the internet requires [object] \\
& write & writing requires [object]
& type & typing requires [object] \\
& cook a curry & cooking a curry requires [object]
& play the piano & playing the piano requires [object] \\
& playing soccer & playing soccer requires [object]
& eating breakfast in bed & eating breakfast in bed requires [object] \\
& paint a house & painting a house requires [object]
& going on a vacation & going on a vacation requires [object] \\
& a goldfish & a goldfish requires [object]
& washing your clothes & washing your clothes requires [object] \\
& cleaning clothing & cleaning clothing requires [object]
& cleaning your room & cleaning your room requires [object] \\

\midrule
\multirow{24}{*}{\rotatebox{90}{\textbf{Element}}}

& water & [object] has water
& news & [object] has news \\
& urine & [object] has urine
& twelve numbers & [object] has twelve numbers \\
& toaster & [object] has toaster
& six strings & [object] has six strings \\
& two wheels & [object] has two wheels
& doorway & doorway has [object] \\
& legs & [object] has legs
& an art show & an art show has [object] \\
& fire & [object] has fire
& ecosystem & ecosystem has [object] \\
& blades & [object] has blades
& foot & [object] has foot \\
& heating system & heating system has [object]
& money & [object] has money \\
& knowledge & [object] has knowledge
& six sides & [object] has six sides \\
& circuit & circuit has [object]
& window frame & window frame has [object] \\
& bathroom & bathroom has [object]
& a document folder & [object] has a document folder \\
& screen & [object] has screen
& keys & [object] has keys \\
& food & [object] has food
& 88 keys & [object] has 88 keys \\
& books & [object] has books
& trash & [object] has trash \\
& tack & [object] has tack
& clothes & [object] has clothes \\
& sofa & sofa has [object]
& computer & computer has [object] \\
& toilet paper & [object] has toilet paper
& air passage & air passage has [object] \\
& rundle & rundle has [object]
& soap & [object] has soap \\
& beer & [object] has beer
& a coat & [object] has a coat \\
& a shower stall & a shower stall has [object]
& table & table has [object] \\
& the movies & the movies have [object]
& clothing & [object] has clothing \\
& bed & bed has [object]
& a wick & [object] has a wick \\
& the date & [object] has the date
& mail & [object] has mail \\
& a cd & [object] has a cd
& & \\

\midrule
\multirow{5}{*}{\rotatebox{90}{\textbf{Material}}}

& wood & [object] is made of wood
& fabric & [object] is made of fabric \\
& leather & [object] is made of leather
& cotton & [object] is made of cotton \\
& metal & [object] is made of metal
& stone & [object] is made of stone \\
& porcelain & [object] is made of porcelain
& plastic & [object] is made of plastic \\
& glass & [object] is made of glass
& paper & [object] is made of paper \\

        \bottomrule
    \end{tabular}
        }
    \caption{OpenScan benchmark attributes of \textit{requirement}, \textit{element}, and \textit{material} aspects.}
\label{dataset_example_3}

\end{table*}

%


\end{document}